\documentclass{ipol}

\usepackage{helvet}
\usepackage{amsmath}
\usepackage{amssymb}
\usepackage{mathtools}
\usepackage{graphicx} 
\usepackage{amsfonts}
\usepackage{subfig}
\usepackage{float}
\usepackage{url}
\usepackage[ruled,vlined]{algorithm2e}
\newtheorem{remark}{Remark}
\DeclareMathAlphabet\mathbfcal{OMS}{cmsy}{b}{n}

\ipolSetTitle{MBO Scheme for Local Chan--Vese Segmentation}
\ipolSetAuthors{Kevin Bui\ipolAuthorMark{1},
                Adina Ciomaga\ipolAuthorMark{2}}
\ipolSetAffiliations{%
\ipolAuthorMark{1} Department of Mathematics, University of California, Irvine, 340 Rowland Hall, Irvine, CA 92697-3875, USA
                   (\texttt{kevinb3@uci.edu})\\
\ipolAuthorMark{2} Université Paris Cité, CNRS, Sorbonne Université, Laboratoire Jacques-Louis Lions (LJLL), F-75006 Paris, France
\&
Octav Mayer Institute of Mathematics, 
Romanian Academy, Ia\c si Branch,
700506 Ia\c si, Romania.
(\texttt{adina@ljll.univ-paris-diderot.fr, adina.ciomaga@acadiasi.ro})}

\begin{document}

\begin{ipolAbstract}
Robust to intensity inhomogeneity, the local Chan--Vese (LCV) model extends the classical Chan--Vese (CV) image segmentation method by incorporating local statistical information around each pixel. Originally, the LCV model was solved using a finite difference scheme, following the approach used for the CV model. As an alternative to the finite difference scheme, a more efficient algorithm based on the Merriman-Bence-Osher (MBO) scheme was later developed for the CV model. In this paper, we derive a similar MBO-based algorithm to solve the LCV model and propose an efficient implementation. The algorithm is developed for both two-phase and multiphase segmentation, and an extension to color images is also discussed. To demonstrate the effectiveness of the proposed approach, we apply it to a variety of grayscale and color images, including medical and microscopy images.
\end{ipolAbstract}

\ipolKeywords{Chan--Vese segmentation, MBO scheme, intensity inhomogeneity, level sets, energy minimization, Mumford-Shah model}

\section{Introduction}\label{sec:intro}

For the past few decades, many algorithms for image segmentation based on variational methods and partial differential equations (PDEs) have been developed. One class of models, including the snakes/active contours, relies on edge-detection functions and evolves the curves toward sharp image gradients \cite{caselles1997geodesic, cohen1991active, kass1988snakes}. However, this class is generally sensitive to noise. For a  demonstration of the active contour model,  see \cite{pierre2014segmentation}.  Another class consists of region-based models, which are more robust to noise as they incorporate both region and boundary information. Mumford and Shah \cite{mumford1989optimal} developed one of the most fundamental region-based models, which approximates images by piecewise-smooth functions. However, the Mumford--Shah model is difficult to solve because it is nonconvex and the set of unknown edges needs to be discretized. 

As a simplification to the Mumford--Shah model, the Chan--Vese model \cite{Chan-Vese-2001} performs two-phase segmentation, where an image has only two regions to segment. 
It was originally solved using a level-set method \cite{osher1988fronts} via an alternating scheme: at each iteration, an Euler--Lagrange equation is first solved using finite differences, and afterward, the constant values for each region are updated by averaging the pixel intensities over the corresponding regions. A detailed explanation and demonstration of the algorithm are presented in \cite{getreuer2012chan}. 
Many other algorithms were later developed to solve the Chan--Vese model. Esedoglu and Tsai \cite{esedog2006threshold} proposed an algorithm based on the Merriman--Bence--Osher (MBO) scheme, a PDE-based method for evolving interfaces by mean curvature \cite{merriman1994motion}. Chan et al. \cite{chan2006algorithms} and Brown et al. \cite{brown2012completely} proposed convex relaxations of the Chan--Vese model, which allow the use of popular convex optimization methods such as ADMM \cite{boyd2011distributed} and the primal--dual hybrid gradient method \cite{chambolle2011first}.

Many variants of the Chan--Vese model have been designed to address different types of images. Vese and Chan \cite{vese2002multiphase} developed a multiphase extension to segment images into $M = 2^m$ regions for some positive integer $m$. Moreover, Chan et al.~\cite{chan2000active} extended the Chan--Vese model to vector-valued images, including RGB images. Li et al.~\cite{li2010multiphase} incorporated fuzzy $c$-means clustering \cite{bezdek1993review} to perform multiphase segmentation with an arbitrary number of regions.

Although the Chan--Vese model is robust to noise and blur in images \cite{Chan-Vese-2001}, it is sensitive to intensity inhomogeneity. Zosso et al.~\cite{zosso2017image} assumed that images with intensity inhomogeneity contain additive illumination bias and modified the Chan--Vese model into a constrained optimization problem. Wang et al.~\cite{wang2010efficient} developed the local Chan--Vese model by adding a local term to the Chan--Vese energy that incorporates statistical information from a neighborhood around each pixel. Their idea is motivated by the observation that pixel intensities within a small neighborhood tend to be more homogeneous than those over the entire image. The simplicity of the local Chan--Vese model has inspired numerous extensions and numerical algorithms \cite{ali2016variational, dong2013new, wang2013efficient}. Most of these methods rely on finite difference schemes. Bui et al.~\cite{bui2019segmentation} designed an alternative MBO-based algorithm to solve the multiphase local Chan--Vese model.

The local Chan--Vese model was originally solved using level-set methods \cite{wang2010efficient}. However, this approach presents several limitations. First, the resulting gradient descent PDE contains degenerate elliptic terms, making it computationally expensive to solve \cite{esedog2006threshold}. Second, while the theoretical formulation involves the Heaviside function, numerical implementations rely on a regularized approximation,  making the algorithm difficult to provide fast segmentation with changing topologies \cite{zosso2017image}. 

To address the issues of the level-set method, we present in this paper an alternative and efficient algorithm for the local Chan--Vese model based on the MBO scheme. This approach was introduced in \cite{bui2019segmentation}, which developed a framework for analyzing scanning tunneling microscopy (STM) images obtained at the California NanoSystems Institute. STM images exhibit varying textures and apparent heights because of the molecular structure and chemical properties of self-assembled monolayers (SAMs). Segmenting these images according to apparent height and texture facilitates the analysis of SAMs and related surfaces. The framework proposed in \cite{bui2019segmentation} consists of three main steps: (i) a cartoon--texture decomposition of the STM image, (ii) segmentation of the cartoon component into regions of homogeneous intensity, and (iii) segmentation of the texture component to separate distinct texture patterns. In this work, we detail the algorithmic implementation of the local Chan--Vese model used in step (ii).
\smallskip

The paper is organized as follows. Section~\ref{sec:background} provides preliminary mathematical background on the Mumford-Shah and Chan--Vese models. Section~\ref{sec:LCV} reviews the local Chan--Vese model and introduces a Ginzburg--Landau approximation to which the MBO scheme can be directly applied. Section~\ref{sec:multiphase} generalizes the local Chan-Vese model to multiphase segmentation, while Section~\ref{sec:vector_value} presents the vector-valued extension.  Section~\ref{sec:experiments} presents experimental results on various grayscale and color images. Lastly, Section~\ref{sec:conclusion} concludes the paper.

\section{Background: Mumford-Shah and Chan--Vese Models} \label{sec:background}

Let $f:\Omega\to\mathbb{R}$ be a given image defined on a domain $\Omega \subset \mathbb{R}^2$, which we assume to be an open rectangle. The goal of image segmentation is to decompose the domain $\Omega$ into several mutually disjoint regions $\{\Omega_i\}_{i=1}^M$ such that
$
\overline{\Omega} = \bigcup_{i=1}^M \overline{\Omega}_i$ and 
$\Omega_j \cap \Omega_k = \emptyset \text{ for all } j \neq k.
$
The image $f$ is assumed to be piecewise smooth: it varies smoothly within each region $\Omega_i$, $i=1,\dots,M$, and exhibits strong discontinuities along the boundaries separating neighboring regions. We denote 
$
\Gamma = \bigcup_{i=1}^M \partial \Omega_i
$ as
the union of all region boundaries in $\Omega$.

The classical Mumford--Shah (MS) segmentation model \cite{mumford1989optimal} is formulated as the minimization of a functional consisting of three terms: an $L^2$ fidelity term, an $H^1$ regularization term enforcing smoothness within each region, and a contour-length regularization term. More precisely, the MS model reads
\begin{align}\tag{MS}
	\label{eq:ms_full}
	\min_{u, \Gamma} \; 
	\lambda \int_{\Omega}(u(x)-f(x))^2 \, dx
	+ \int_{\Omega\setminus \Gamma}|\nabla u(x)|^2 \, dx
	+ \nu |\Gamma|,
\end{align}
where $u:\Omega \rightarrow \mathbb{R}$ is a piecewise-smooth approximation of the image $f$, $|\Gamma|$ denotes the total length of the set of discontinuities, and $\lambda,\nu>0$ are weight parameters for the fidelity and perimeter terms, respectively.

To simplify the model, the function $u$ may be restricted to be piecewise constant, namely
\[
u(x) = \sum_{i=1}^M c_i \chi_{\Omega_i}(x),
\]
where $c_i\in\mathbb{R}$ denotes the constant intensity value of region $\Omega_i$ for each $i=1,\dots,M$ and the characteristic function $\chi_A$ for a set $A$ is given by
\[
\chi_{A}(x) =
\begin{cases}
1, & x \in A, \\
0, & x \notin A.
\end{cases}
\]
In this setting, the smoothing term over $\Omega\setminus\Gamma$ vanishes, and the regularization reduces to a perimeter penalization. Without loss of generality, the weight parameter $\nu$ can be fixed to $\nu=1$.

In the particular case $M=2$, this formulation leads to the Chan--Vese (CV) model \cite{Chan-Vese-2001}
\begin{align}\label{eq:cv_model}
	\min_{c_1, c_2, \Gamma}
	\lambda \int_{\Omega_1}(f(x)-c_1)^2 \, dx
	+ \lambda \int_{\Omega_2}(f(x)-c_2)^2 \, dx
	+ |\Gamma|.
\end{align}
This minimization problem can be reformulated using the level-set method. Let $\phi:\Omega \rightarrow \mathbb{R}$ be a Lipschitz continuous level-set function such that
\[
\Gamma = \{x \in \Omega : \phi(x) = 0\}, \qquad
\Omega_1 = \{x \in \Omega : \phi(x) > 0\}, \qquad
\Omega_2 = \{x \in \Omega : \phi(x) < 0\}.
\]
Let $H:\mathbb{R} \to\{0,1\}$ denote the Heaviside function, i.e., $H = \chi_{[0,\infty)}.$
Then 
\[ 
H(\phi(x)) = 1 \; \text{ for } x \in \Omega_1, \quad 
H(\phi(x)) = 0 \; \text{ for } x \in \Omega_2. 
\]
Introducing the functionals
\begin{align*}
\mathcal{E}_{\mathrm{fid}}(\chi,\mathbf{c};f)
&\coloneqq \int_{\Omega} (f(x)-c_1)^2 \chi(x)
+ (f(x)-c_2)^2 (1-\chi(x)) \, dx, \\
\mathcal{E}_{\mathrm{per}}(\chi)
&\coloneqq \int_{\Omega} |\nabla \chi(x)| \, dx,
\end{align*}
where $\chi \in BV(\Omega)$ and $\mathbf{c}=(c_1,c_2)\in\mathbb{R}^2$, the level-set formulation of \eqref{eq:cv_model} can be written as
\begin{align}\tag{CV}\label{eq:CV}
\min_{\phi,\mathbf{c}}
\; \lambda \, \mathcal{E}_{\mathrm{fid}}(H(\phi),\mathbf{c};f)
+ \mathcal{E}_{\mathrm{per}}(H(\phi)).
\end{align}

\section{Local Chan--Vese Model and the MBO Scheme} \label{sec:LCV}

Real images often exhibit illumination bias, typically manifested as smooth,
artificial intensity variations across the spatial domain.
Since the CV model assumes piecewise-constant intensity
values within each region, it may perform poorly in the presence of such
inhomogeneities because the intensity distributions of neighboring regions can
overlap, leading to blurred interfaces and inaccurate segmentations.

To enhance robustness with respect to illumination bias, Wang et al.~\cite{wang2010efficient}
proposed augmenting the CV model with a local fitting term based on linear,
high-pass filtering.
Let $g_k$ denote a smoothing kernel (e.g.,\ a mean or Gaussian kernel) with a
$(k\times k)$ window.
The convolution $g_k\ast f$ acts as a low-pass filter, preserving large-scale
intensity variations while attenuating fine-scale structures.
Consequently, the difference $f-g_k\ast f$ emphasizes edges and local contrast,
thereby reducing the influence of slowly varying intensity distortions.
Accordingly, the local energy term is defined as
\begin{align}
\mathcal{E}_{\mathrm{loc}}(\chi,\mathbf{d};f)
=
\mathcal{E}_{\mathrm{fid}}(\chi,\mathbf{d}; g_k \ast f - f),
\end{align}
where $\mathbf{d}=(d_1,d_2)\in\mathbb{R}^2$ denotes region-wise constants\footnote{We use $g_k\ast f - f$ instead of $f-g_k\ast f$ to remain consistent
with the notation in \cite{wang2010efficient}. This sign convention does not
affect the removal of illumination bias or the localization of region
boundaries.}.

The local Chan--Vese (LCV) model \cite{wang2010efficient} is then formulated as
\begin{align}\tag{LCV}\label{eq:LCV}
\min_{\phi,\mathbf{c},\mathbf{d}} \;
\lambda \, \mathcal{E}_{\mathrm{fid}}(H(\phi),\mathbf{c};f)
+
\beta \, \mathcal{E}_{\mathrm{loc}}(H(\phi),\mathbf{d};f)
+
\mathcal{E}_{\mathrm{per}}(H(\phi)),
\end{align}
where $\beta\ge0$ controls the relative influence of the local term.
When $\beta=0$, the model reduces to the classical CV formulation.

\begin{remark}[Interpretation as a vector-valued CV model]
\label{rem:LCV_vector_interpretation}
The LCV functional \eqref{eq:LCV} can be interpreted as a particular instance of
the vector-valued CV model \cite{chan2000active} applied to a two-channel image, where the two channels are the original image $f$ and the weighted image difference $\displaystyle \sqrt{\frac{\beta}{\lambda}}(g_k\ast f -f )$.
Hence, the fidelity and local fitting terms correspond to the
contributions of the two channels in the vector-valued CV model.
This viewpoint emphasizes that the local term does not introduce a fundamentally
new variational structure but rather enriches the data representation by
combining the original image with its linearly filtered version.
Since $g_k\ast f-f$ corresponds to a frequency-weighted transformation of $f$,
the LCV model can be viewed as exploiting complementary spectral information,
which improves the discrimination between regions.
The general vector-valued formulation and its numerical treatment are discussed
in Section~\ref{sec:vector_value}.
\end{remark}

\subsection{Ginzburg--Landau Approximation}\label{subsec:GL-approx}

Esedoglu and Tsai \cite{esedog2006threshold} proposed an efficient method to minimize the two-phase energy functional \eqref{eq:CV}. Their algorithm is based on the MBO scheme \cite{merriman1994motion}, a threshold dynamics method for evolving interfaces by mean curvature. The main idea is to use a smooth approximation of the Heaviside function $H$ and to replace the contour-length term $\mathcal{E}_{\text{per}}(H(\phi))$ in \eqref{eq:CV} by a Ginzburg--Landau functional
\begin{align} \label{eq:GL}
     \mathcal{E}_{\text{GL}}^{\varepsilon}(\phi)
     =
     \int_{\Omega}
     \left(
     \frac{\varepsilon}{2}|\nabla \phi(x)|^2
     + \frac{2}{\varepsilon} W(\phi(x))
     \right)
     \, dx,
\end{align}
where $W(s)=s^2(1-s)^2$ is a double-well potential. It was shown by Modica \cite{modica1987gradient} that the Ginzburg--Landau functional $\mathcal{E}_{\text{GL}}^{\varepsilon}$ $\Gamma$-converges to the perimeter functional $\mathcal{E}_{\text{per}}(H(\phi))$ as $\varepsilon \to 0^+$.

Parametrizing the descent direction by an artificial time variable $t>0$,
the Euler--Lagrange equation associated with \eqref{eq:GL} yields the
Allen--Cahn equation (also known as the phase-field equation)
\begin{align} \label{eq:allen_cahn}
\partial_t\phi (x,t)
=
\varepsilon \Delta \phi(x,t)
-
\frac{2}{\varepsilon} W'(\phi(x,t)),
\end{align}
with initial data $\phi(x,0)$ corresponding to the initial placement of the contour. To solve \eqref{eq:allen_cahn}, the MBO scheme can be applied, which consists of first solving a linear heat equation followed by  thresholding. Thresholding is justified by the fact that the nonlinear ordinary differential equation
\begin{equation}\label{eq:ODE-W}
\partial_t\phi(x,t)
=
-
\frac{2}{\varepsilon} W'(\phi(x,t))
\end{equation}
admits two stable stationary solutions at $\bar{\phi}\equiv 0$ and $\bar{\phi}\equiv 1$ and an unstable stationary solution at $\tilde{\phi}\equiv \frac{1}{2}$. The key observation underlying the MBO scheme is that, as $\varepsilon \to 0$, for any fixed time $t>0$, the solution $\phi$ converges to one of the two stable equilibrium values, depending on the basin of attraction in which it is initialized.

To incorporate the MBO scheme in solving the LCV minimization problem
\eqref{eq:LCV}, we consider the following diffuse-interface approximation:
\begin{align}\tag{eLCV}\label{eq:GL_LCV}
\min_{u,\mathbf{c},\mathbf{d}}
\;
\lambda \mathcal{E}_{\text{fid}}(u,\mathbf{c})
+ \beta \mathcal{E}_{\text{loc}}(u,\mathbf{d})
+ \mathcal{E}_{\text{GL}}^{\varepsilon}(u).
\end{align}
At the diffuse-interface level, we replace the level-set function $\phi$ by a phase-field variable $u:\Omega\to[0,1]$.
The function $u$ provides a smooth approximation of the characteristic
function of one phase, taking values close to $1$ inside the region and close
to $0$ outside. In this formulation, the sharp interface represented by the
zero level set of $\phi$ is replaced by a smooth interface 
governed by the Ginzburg--Landau energy \cite[Theorem 1]{modica1987gradient}.
Accordingly, the Heaviside function $H(\phi)$ appearing in the original LCV
model is approximated directly by the phase-field variable $u$.

To solve \eqref{eq:GL_LCV}, we adopt an alternating minimization strategy. We introduce an additional variable $t \ge 0$, interpreted as an artificial time parameter used to describe the gradient-flow dynamics associated with the diffuse energy; steady states of this evolution correspond
to critical points of the diffuse-interface functional. The initialization of $u$ will be discussed later.

\begin{itemize}
\item Keeping $u$ fixed and minimizing with respect to $\mathbf{c}$ and
$\mathbf{d}$ yields
\begin{align}
c_1 &= \mathcal{A}[u;f], &
c_2 &= \mathcal{A}[1-u;f], \label{eq:c_update} \\
d_1 &= \mathcal{A}[u;g_k\ast f-f], &
d_2 &= \mathcal{A}[1-u;g_k\ast f-f]. \label{eq:d_update}
\end{align}
where the averaging operator is given by

$$\mathcal{A}[u;f](t):= \frac{
\displaystyle \int_{\Omega} u(x,t) f(x)\,dx 
\;}{\; 
\displaystyle  \int_{\Omega} u(x,t)\,dx}.$$

\item Keeping $\mathbf{c}$ and $\mathbf{d}$ fixed, calculus of variations
leads to the following Euler--Lagrange equation satisfied by $u$ in $\Omega$:
\begin{align}\tag{EL}\label{eq:Euler_Lagrange1}
\partial_t u
=
\varepsilon \Delta u
-
\frac{2}{\varepsilon} W'(u)
-
\lambda \mathcal{L}[\mathbf{c};f]
-
\beta \mathcal{L}[\mathbf{d}; g_k \ast f - f],
\end{align}
where the spatial operator $\mathcal{L}$ is defined by

\begin{align*}
\mathcal{L}[\mathbf{c};f](x)
:=
(f(x)-c_1)^2 - (f(x)-c_2)^2.
\end{align*}
\end{itemize}

\subsection{The MBO Scheme for the LCV Model}\label{subsec:MBO-LCV}

We propose an MBO-type scheme to solve the Euler–Lagrange equation \eqref{eq:Euler_Lagrange1}, which computes approximations of $u$ at discrete time instances. The spatial discretization is then chosen as a discrete analogue of the corresponding continuous formulation. 

\subsubsection{Temporal discretization}

We numerically solve the Euler--Lagrange equation using an operator-splitting approach that decomposes the evolution into three simpler substeps. At each iteration, we first solve a linear ordinary differential equation (ODE) using a finite difference scheme, then solve a heat equation under periodic boundary conditions, and finally handle the nonlinear ODE induced by the double-well potential through a thresholding operation.

Let $dt>0$ be a fixed time step. Starting from an initial datum $u^0$, we iteratively construct a sequence of time-discrete, space-continuous functions
$\{w^{n}, v^{n}, u^n\}_{n\geq 1}$ as follows.

\begin{itemize}
\item \textbf{Step 1 (ODE).}
Let $w^{n+1}$ be the discrete solution obtained from $u^n$ by applying a forward finite difference scheme to the linear ODE component of \eqref{eq:Euler_Lagrange1}:
\begin{align}\label{eq:step1}
\frac{w^{n+1}-u^n}{dt}
=
-\lambda \mathcal{L}[\mathbf{c};f]
-\beta \mathcal{L}[\mathbf{d}; g_k \ast f - f].
\end{align}

\item \textbf{Step 2 (Diffusion).}
Let $v^{n+1}$ be the solution of the heat equation with periodic boundary conditions and initial data $w^{n+1}$. The solution at time $dt$ admits the closed-form representation
\begin{align}\label{eq:step2}
v^{n+1}
=
\mathcal{F}^{-1}
\left\{
\frac{1}{1+dt|\xi|^2}
\mathcal{F}(w^{n+1})(\xi)
\right\},
\end{align}
where $\mathcal{F}$ and $\mathcal{F}^{-1}$ denote the two-dimensional Fourier transform and its inverse.

\item \textbf{Step 3 (Thresholding).}
Finally, $u^{n+1}$ is obtained by solving the nonlinear ODE associated with the double-well potential via a thresholding operation:
\begin{align}\label{eq:step3}
u^{n+1}
=
\begin{cases}
0, & \text{if } v^{n+1} \le \tfrac{1}{2}, \\
1, & \text{if } v^{n+1} > \tfrac{1}{2}.
\end{cases}
\end{align}
\end{itemize}
Altogether, this procedure yields an iterative algorithm for solving \eqref{eq:LCV}. At each iteration, the auxiliary variables $\mathbf{c}$ and $\mathbf{d}$ are updated according to \eqref{eq:c_update} and \eqref{eq:d_update}, after which the Euler--Lagrange equation is advanced using Steps 1-3, given by ~\eqref{eq:step1}--\eqref{eq:step3}.

For initialization, an effective choice is the characteristic function
\[
u^0(x)
=
\chi_{\left\{
\sin\!\left(\frac{\pi x_1}{5}\right)
\sin\!\left(\frac{\pi x_2}{5}\right) > 0
\right\}}(x),
\]
corresponding to a separable oscillatory pattern, as suggested in \cite{getreuer2012chan}.

The overall algorithm for minimizing \eqref{eq:LCV} is summarized in Algorithm~\ref{alg:twophase}.

\begin{algorithm}[t!]
\SetAlgoLined \smallskip
\textbf{Input:}
Image $f$;
model parameters $\lambda, \beta$;
time-step parameter $dt$; tolerance \texttt{tol}.\\
\textbf{Initialize:}
\[
u^0
=
\chi_{\left\{
\sin \left( \frac{\pi x_1}{5} \right)
\sin\left(\frac{\pi x_2}{5} \right)>0
\right\}}.
\]

\For{$n=1$ to $N$}{\smallskip

Update $\mathbf{c}^{n+1} \coloneqq (c^{n+1}_1, c^{n+1}_2)$  and $\mathbf{d}^{n+1} \coloneqq (d^{n+1}_1, d^{n+1}_2)$ by
\begin{align*}
&c_1^{n+1} = \mathcal A[u^n; f], 
\qquad 
c_2^{n+1} = \mathcal A[1-u^n; f], \\[0.3em]
&d_1^{n+1} = \mathcal A[u^n; g_k\ast f - f], 
\qquad 
d_2^{n+1} = \mathcal A[1-u^n; g_k\ast f - f].
\end{align*}

Compute $u^{n+1}$:
\begin{align*}
& w^{n+1}
= u^n
- dt \Big(
\lambda \mathcal L[\mathbf{c}^{n+1}; f]
+ \beta \mathcal L[\mathbf{d}^{n+1}; g_k\ast f - f]
\Big), \\
& v^{n+1}
= \mathcal{F}^{-1}
\left\{
\frac{1}{1+dt|\xi|^2}\,
\mathcal{F}(w^{n+1})(\xi)
\right\}, \\
& u^{n+1}
=
\begin{cases}
0, & \text{if } v^{n+1} \le \tfrac{1}{2}, \\
1, & \text{if } v^{n+1} > \tfrac{1}{2}.
\end{cases}
\end{align*}

\If{$\|u^{n+1} - u^n\|_2 < \texttt{tol}$}{
break
}
}

\textbf{Output:} segmented image $u^{n+1}$.\\
\caption{MBO algorithm for two-phase LCV segmentation}
\label{alg:twophase}
\end{algorithm}

\subsubsection{Spatial Discretization}
Although Algorithm \ref{alg:twophase} is formulated in a time-discrete, space-continuous setting, its numerical implementation relies on a spectral discretization in space. All spatially dependent quantities are sampled on a uniform Cartesian grid, and spatial derivatives are computed in the Fourier domain. Below, we describe which parts of the algorithm are spatially discretized:

\begin{itemize}
\item The diffusion step in Step 2, provided by \eqref{eq:step2}, is implemented by applying the discrete Fourier transform to $w^{n+1}$, multiplying each Fourier mode by the transfer function $(1+dt|\xi|^2)^{-1}$, and transforming it back to physical space. This corresponds to a spectral discretization of the Laplacian operator with periodic boundary conditions. In contrast to classical finite-difference discretizations of the Laplacian, which lead to cosine-based symbols of the form $2(\cos(h\xi)-1)/h^2$, the present approach preserves the continuous Laplacian spectrum up to the Nyquist frequency. As a result, the scheme benefits from high accuracy for smooth solutions and avoids grid-induced anisotropy.
\item The ODE and thresholding steps are evaluated pointwise in physical space on the same grid, yielding a fully discrete algorithm consistent with the underlying space-continuous formulation.
\item The updates of the auxiliary variables $\mathbf{c}$ and $\mathbf{d}$ involve spatial integrals over the image domain. In practice, these integrals are approximated by finite sums over the grid. For a uniform grid with spacing $h>0$, the functional $\mathcal{A}[u;f]$ is approximated by
\begin{align*}
\mathcal{A}_h[u;f]
:= \frac{\displaystyle\sum_{i,j} u_{i,j} f_{i,j}}{\displaystyle\sum_{i,j} u_{i,j}},
\end{align*}
where $(i,j)$ index the grid points and $u_{i,j}, f_{i,j}$ are the sampled values of $u,f$ on the grid points, respectively.
\end{itemize}

\begin{remark}
The numerical strategy adopted in this work follows a ``minimize, then discretize" approach. Starting from the continuous formulation of the energy functional, we first derive the associated Euler--Lagrange gradient-flow equation. This evolution equation is then discretized in time and space using a spectral method combined with a thresholding step. Owing to the linearity of the diffusion operator and the use of the Fourier transform, the resulting scheme admits an explicit solution with a fast FFT-based solver and good control over stability properties.
The motivation for this strategy is to achieve accurate segmentation results with precise boundary localization. Since discretization is performed at the PDE level, the scheme remains variationally consistent: as the time step $dt$ is refined and the number of iterations increases, the discrete evolution approximates the continuous minimizer. As shown in Section~\ref{sec:experiments}, the numerical evolution is smooth, the results are stable, and the energy decreases monotonically.
By contrast, the ``discretize, then minimize approach" first discretizes the energy functional and then applies numerical optimization methods such as gradient descent. While this strategy often yields fast convergence, it may suffer from grid-dependent effects and reduced consistency with respect to the underlying continuous model, particularly when accurate geometric localization of interfaces is required.
\end{remark}

\section{Multiphase Segmentation}\label{sec:multiphase}

The LCV model and its numerical scheme can be naturally extended to the multiphase case. In this setting, the image domain $\Omega$ is assumed to be partitioned into a collection of $M$ mutually disjoint regions $\{\Omega_i\}_{i=1}^{M}$, where $M=2^m$ for some positive integer $m$. Each region $\Omega_i$ is determined by a vector-valued level-set function $\Phi=(\phi_1,\ldots,\phi_m)$. 

Each index $i\in\{1,\ldots,2^m\}$ can be written in binary form as
$ \displaystyle
i = \sum_{j=1}^{m} a_j 2^{j-1} + 1$, where $a_j \in \{0,1\}.
$
Accordingly, for each $j=1,\ldots,m$, we define
\begin{align*}
\Omega_{j,a_j} =
\begin{cases}
\{x\in\Omega : \phi_j(x)>0\}, & a_j=0,\\
\{x\in\Omega : \phi_j(x)<0\}, & a_j=1.
\end{cases}
\end{align*}
Each region $\Omega_i$ is then given by
$\displaystyle
\Omega_i = \bigcap_{j=1}^{m} \Omega_{j,a_j}.
$

For example, when $M=4$ $(m=2)$, the four regions are
\begin{align*}
\Omega_1 &= \{x\in\Omega : \phi_1(x)>0,\;\phi_2(x)>0\}, 
\Omega_2 = \{x\in\Omega : \phi_1(x)<0,\;\phi_2(x)>0\}, \\
\Omega_3 &= \{x\in\Omega : \phi_1(x)>0,\;\phi_2(x)<0\}, 
\Omega_4 = \{x\in\Omega : \phi_1(x)<0,\;\phi_2(x)<0\}.
\end{align*}

Let $\mathbf{H}:\mathbb{R}^m\to\mathbb{R}^m$ be the vector-valued Heaviside function given by 
$$\mathbf{H}(\Phi)=(H(\phi_1),\ldots,H(\phi_m)). $$
We denote $\chi_i$ as the characteristic function of $\Omega_i$ for $i=1,\ldots,M$.
For $\mathbf{c}=(c_1,\ldots,c_M)\in\mathbb{R}^M$ and
$\mathbf{d}=(d_1,\ldots,d_M)\in\mathbb{R}^M$,
the multiphase local Chan--Vese energy functional is defined as
\begin{align}\tag{M-LCV}\label{eq:M-LCV}
\min_{\Phi,\mathbf{c},\mathbf{d}}
\;
\lambda \mathcal{E}_{\mathrm{fid}}^M(\mathbf{H}(\Phi),\mathbf{c};f)
+
\beta \mathcal{E}_{\mathrm{loc}}^M(\mathbf{H}(\Phi),\mathbf{d};f)
+
\mathcal{E}_{\mathrm{per}}^m(\mathbf{H}(\Phi)),
\end{align}
where
\begin{align*}
\mathcal{E}_{\mathrm{fid}}^M(\mathbf{H}(\Phi),\mathbf{c};f)
&=
\sum_{i=1}^{M}
\int_{\Omega}(f(x)-c_i)^2 \chi_i(x)\,dx,\\
\mathcal{E}_{\mathrm{loc}}^M(\mathbf{H}(\Phi),\mathbf{d};f)
&=
\mathcal{E}_{\mathrm{fid}}^M(\mathbf{H}(\Phi),\mathbf{d}; g_k\ast f - f),\\
\mathcal{E}_{\mathrm{per}}^m(\mathbf{H}(\Phi))
&=
\sum_{j=1}^{m}
\int_{\Omega} |\nabla H(\phi_j(x))|\,dx.
\end{align*}

\begin{remark}
The characteristic functions $\chi_i$ can be explicitly written in terms of $\mathbf{H}(\Phi)$. For instance, when $M=2$ $(m=1)$, we recover the two-phase LCV model with
\[
\chi_1 = H(\phi_1), \qquad \chi_2 = 1-H(\phi_1).
\]
For $M=4$ $(m=2)$, the four-phase case is given by
\begin{alignat*}{2}
\chi_1 &= H(\phi_1)H(\phi_2), \qquad &
\chi_2 &= (1-H(\phi_1))H(\phi_2),\\
\chi_3 &= H(\phi_1)(1-H(\phi_2)), \qquad &
\chi_4 &= (1-H(\phi_1))(1-H(\phi_2)).
\end{alignat*}
For completeness, we also give the explicit expressions for the characteristic
functions in the eight-phase case $(M=8,\, m=3)$:
\begin{alignat*}{2}
\chi_1 &= H(\phi_1)H(\phi_2)H(\phi_3), \qquad
&& \chi_2 = (1-H(\phi_1))H(\phi_2)H(\phi_3),\\
\chi_3 &= H(\phi_1)(1-H(\phi_2))H(\phi_3), \qquad
&& \chi_4 = (1-H(\phi_1))(1-H(\phi_2))H(\phi_3),\\
\chi_5 &= H(\phi_1)H(\phi_2)(1-H(\phi_3)), \qquad
&& \chi_6 = (1-H(\phi_1))H(\phi_2)(1-H(\phi_3)),\\
\chi_7 &= H(\phi_1)(1-H(\phi_2))(1-H(\phi_3)), \qquad
&& \chi_8 = (1-H(\phi_1))(1-H(\phi_2))(1-H(\phi_3)).
\end{alignat*}
\end{remark}

\subsection{Ginzburg--Landau Approximation for Multiphase LCV}
To apply the MBO scheme to the multiphase LCV model, we introduce a diffuse-interface approximation based on the Ginzburg--Landau functional. As in the two-phase setting, the perimeter regularization is replaced by a phase-field formulation that allows interfaces to be represented implicitly through smooth variables. This approximation leads to a phase-field representation in which each level-set component is associated with its own Ginzburg--Landau energy, while coupling between phases is retained through the fidelity and local data terms. This formulation provides a natural framework for extending the MBO splitting strategy to multiphase segmentation.

Let $\mathbf{u}=(u_1,\ldots,u_m):\Omega \rightarrow [0,1]^m$. The multiphase LCV model can be approximated by the following Ginzburg--Landau energy:

\begin{align}\tag{eM-LCV}\label{eq:eM-LCV}
\min_{\mathbf{u},\mathbf{c},\mathbf{d}}
\;
\lambda \mathcal{E}_{\mathrm{fid}}^M(\mathbf{u},\mathbf{c};f)
+
\beta \mathcal{E}_{\mathrm{loc}}^M(\mathbf{u},\mathbf{d};f)
+
\mathcal{E}_{\mathrm{GL}}^{m,\varepsilon}(\mathbf{u}),
\end{align}
where the perimeter term is approximated by
\begin{align}
\mathcal{E}_{\mathrm{GL}}^{m,\varepsilon}(\mathbf{u})
=
\sum_{i=1}^{m}
\int_{\Omega}
\frac{\varepsilon}{2}|\nabla u_i(x)|^2
+
\frac{2}{\varepsilon} W(u_i(x))
\,dx.
\end{align}

\subsection{The MBO Scheme for Multiphase LCV}

The application of the MBO scheme to \eqref{eq:eM-LCV} follows naturally from the
two-phase formulation. 
Because of the additive structure of the Ginzburg--Landau approximation, the
Euler--Lagrange equations associated with $\mathcal{E}_{\mathrm{GL}}^{m,\varepsilon}$ yield a system of $m$ evolution equations, one for each phase-field variable $u_i$, which are coupled through the fidelity and local data terms.
As before, the artificial time variable  $t>0$ is introduced solely to derive the Euler–Lagrange gradient-flow equation associated with the diffuse energy.

For clarity, we detail the derivation in the four-phase case $(m=2)$.
To simplify notation, we introduce the averaging operator
\begin{align*}
\mathcal{A}[u_1,u_2;f](t)
:=
\frac{\displaystyle\int_{\Omega} f(x)\,u_1(x,t)\,u_2(x,t)\,dx}
{\displaystyle\int_{\Omega} u_1(x,t)\,u_2(x,t)\,dx}
\end{align*}
and the spatial operator
\begin{align*}
\mathcal{L}(\mathbf{c},u;f)(x,t)
:=
(f(x)-c_1)^2 u(x,t) - 
(f(x)-c_2)^2 u(x,t) +\\ 
(f(x)-c_3)^2(1-u(x,t)) - 
(f(x)-c_4)^2(1-u(x,t)).
\end{align*}

The updates for the region averages $\mathbf{c}=(c_1,\dots,c_4)$ and
$\mathbf{d}=(d_1,\dots,d_4)$ are then given by
\begin{align*}
c_1 &= \mathcal{A}[u_1,u_2;f], &
d_1 &= \mathcal{A}[u_1,u_2;g_k\ast f-f],\\
c_2 &= \mathcal{A}[1-u_1,u_2;f], &
d_2 &= \mathcal{A}[1-u_1,u_2;g_k\ast f-f],\\
c_3 &= \mathcal{A}[u_1,1-u_2;f], &
d_3 &= \mathcal{A}[u_1,1-u_2;g_k\ast f-f],\\
c_4 &= \mathcal{A}[1-u_1,1-u_2;f], &
d_4 &= \mathcal{A}[1-u_1,1-u_2;g_k\ast f-f].
\end{align*}

By calculus of variations, the Euler--Lagrange equations for $u_1$ and $u_2$
are given by
\begin{align}\label{eq:M-EL1}
\partial_t u_1
&=
\varepsilon \Delta u_1
- \frac{2}{\varepsilon} W'(u_1)
- \lambda \mathcal{L}(\mathbf{c},u_2;f)
- \beta \mathcal{L}(\mathbf{d},u_2;g_k\ast f-f), \\
\label{eq:M-EL2}
\partial_t u_2
&=
\varepsilon \Delta u_2
- \frac{2}{\varepsilon} W'(u_2)
- \lambda \mathcal{L}(\bar{\mathbf{c}},u_1;f)
- \beta \mathcal{L}(\bar{\mathbf{d}},u_1;g_k\ast f-f).
\end{align}
where $\bar{\mathbf{c}}=(c_1,c_3,c_2,c_4)$ and
$\bar{\mathbf{d}}=(d_1,d_3,d_2,d_4)$.

Each equation is solved using the same MBO splitting strategy as in the
two-phase case, consisting of a linear ODE step, a diffusion step, and a
thresholding operation. For initialization, we take
\begin{align*}
u_1^1(x,0) &= \chi_{\{\sin(\frac{\pi x_1}{3})\sin(\frac{\pi x_2}{3})>0\}}(x), \\
u_2^1(x,0) &= \chi_{\{\sin(\frac{\pi x_1}{10})\sin(\frac{\pi x_2}{10})>0\}}(x),
\end{align*}
as suggested in \cite{bui2019segmentation}.

The algorithm to solve the four-phase LCV model is summarized in
Algorithm~\ref{alg:fourphase}.

\begin{algorithm}[h!]
\SetAlgoLined
\medskip
\textbf{Input:} Image $f$; model parameters $\lambda, \beta$;
time-step parameter $dt$; tolerance \texttt{tol}. 
\medskip\\
\textbf{Initialize} $\mathbf{u}^0 \coloneqq (u_1^0, u_2^0)$:
    \begin{align*}
    u_1^0 &= \chi_{\{\sin(\frac{\pi x_1}{3})\sin(\frac{\pi x_2}{3}) > 0\}}, \\
    u_2^0 &= \chi_{\{\sin(\frac{\pi x_1}{10})\sin(\frac{\pi x_2}{10}) > 0\}}.
    \end{align*}
\For{$n=1$ to $N$}{
    \smallskip
    Update 
    $\displaystyle \mathbf{c}^{n+1} \coloneqq (c_1^{n+1}, c_2^{n+1}, c_3^{n+1}, c_4^{n+1})$
    and 
    $\displaystyle \bar{\mathbf{c}}^{n+1} \coloneqq (c_1^{n+1}, c_3^{n+1}, c_2^{n+1}, c_4^{n+1})$:
    \begin{align*}
    c_1^{n+1} &= \mathcal{A}[u_1^{n}, u_2^{n}; f], \\
    c_2^{n+1} &= \mathcal{A}[1-u_1^{n}, u_2^{n}; f], \\
    c_3^{n+1} &= \mathcal{A}[u_1^{n}, 1-u_2^{n}; f], \\
    c_4^{n+1} &= \mathcal{A}[1-u_1^{n}, 1-u_2^{n}; f].
    \end{align*}
    \smallskip
    Update $\mathbf{d}^{n+1} \coloneqq (d_1^{n+1}, d_2^{n+1}, d_3^{n+1}, d_4^{n+1})$
    and $\bar{\mathbf{d}}^{n+1} \coloneqq (d_1^{n+1}, d_3^{n+1}, d_2^{n+1}, d_4^{n+1})$:
    \begin{align*}
    d_1^{n+1} &= \mathcal{A}[u_1^{n}, u_2^{n}; g_k \ast f - f], \\
    d_2^{n+1} &= \mathcal{A}[1-u_1^{n}, u_2^{n}; g_k \ast f - f], \\
    d_3^{n+1} &= \mathcal{A}[u_1^{n}, 1-u_2^{n}; g_k \ast f - f], \\
    d_4^{n+1} &= \mathcal{A}[1-u_1^{n}, 1-u_2^{n}; g_k \ast f - f].
    \end{align*}
    \smallskip
    Compute $\mathbf{u}^{n+1} = (u_1^{n+1}, u_2^{n+1})$:
    \begin{align*}
    w_1^{n+1} &= u_1^{n}
    - dt \Big(
    \lambda \mathcal{L}(\mathbf{c}^{n+1}, u_2^{n}; f)
    + \beta \mathcal{L}(\mathbf{d}^{n+1}, u_2^{n}; g_k \ast f - f)
    \Big), \\
    v_1^{n+1} &=
    \mathcal{F}^{-1}
    \left\{
    \frac{1}{1+dt|\xi|^2}\mathcal{F}(w_1^{n+1})(\xi)
    \right\}, \\
    u_1^{n+1} &=
    \begin{cases}
    0, & v_1^{n+1} \le \tfrac{1}{2}, \\
    1, & v_1^{n+1} > \tfrac{1}{2},
    \end{cases}
    \\[0.8em]
    w_2^{n+1} &= u_2^{n}
    - dt \Big(
    \lambda \mathcal{L}(\bar{\mathbf{c}}^{n+1}, u_1^{n+1};f)
    + \beta \mathcal{L}(\bar{\mathbf{d}}^{n+1}, u_1^{n+1};g_k \ast f - f)
    \Big), \\
    v_2^{n+1} &=
    \mathcal{F}^{-1}
    \left\{
    \frac{1}{1+dt|\xi|^2}\mathcal{F}(w_2^{n+1})(\xi)
    \right\}, \\
    u_2^{n+1} &= 
    \begin{cases}
    0, & v_2^{n+1} \le \tfrac{1}{2}, \\
    1, & v_2^{n+1} > \tfrac{1}{2}.
    \end{cases}
    \end{align*}
\If{$\|\mathbf{u}^{n+1} - \mathbf{u}^{n}\|_2 < \texttt{tol}$}{
    break
}
}
\textbf{Output:} segmented phases $\mathbf{u}^{n+1}$.\\
\caption{MBO algorithm for four-phase LCV segmentation}
\label{alg:fourphase}
\end{algorithm}

\begin{remark}
The MBO scheme extends naturally to the general $M$-phase LCV model.
Following the four-phase derivation, the operators $\mathcal{L}$ and
$\mathcal{A}$ are generalized by considering all product combinations
$\prod_{i=1}^{m} v_i$, where $v_i \in \{u_i,\,1-u_i\}$ for each
$i=1,\ldots,m$.
\end{remark}

\section{Vector-Valued Extension} \label{sec:vector_value}

The $M$-phase LCV model extends naturally to vector-valued images.
Let $\mathbf{f}=(f_1,\ldots,f_C) : \Omega \to \mathbb{R}^C$ denote a multichannel image with
$C$ channels. The convolution kernel $g_k$ acts channel-wise, namely
\[
g_k \ast \mathbf{f} := (g_k \ast f_1,\ldots,g_k \ast f_C).
\]
We introduce region-dependent constants
\[
\vec{\mathbf{c}} = (\mathbf{c}_1,\ldots,\mathbf{c}_M), 
\qquad
\vec{\mathbf{d}} = (\mathbf{d}_1,\ldots,\mathbf{d}_M)
\in \mathbb{R}^{M\times C},
\]
where $\mathbf{c}_i=(c_{i,1},\ldots,c_{i,C})$ and
$\mathbf{d}_i=(d_{i,1},\ldots,d_{i,C})$ represent the mean intensity
and local contrast vectors associated with region $\Omega_i$. The multichannel fidelity and local fitting energies are defined by
\begin{align}
\mathcal{E}_{\mathrm{fid}}^{M,\mathrm{vec}}(\mathbf{H}(\Phi),\vec{\mathbf{c}})
&=
\sum_{i=1}^M
\int_\Omega
\|\mathbf{f}(x)-\mathbf{c}_i\|_2^2\,\chi_i(x)\,dx
\nonumber\\
&=
\sum_{i=1}^M
\int_\Omega
\sum_{j=1}^C (f_j(x)-c_{i,j})^2\,\chi_i(x)\,dx,
\label{eq:mc_fid}
\\
\mathcal{E}_{\mathrm{loc}}^{M,\mathrm{vec}}(\mathbf{H}(\Phi),\vec{\mathbf{d}})
&=
\sum_{i=1}^M
\int_\Omega
\|g_k\ast\mathbf{f}(x)-\mathbf{f}(x)-\mathbf{d}_i\|_2^2\,\chi_i(x)\,dx
\nonumber\\
&=
\sum_{i=1}^M
\int_\Omega
\sum_{j=1}^C
\bigl(g_k\ast f_j(x)-f_j(x)-d_{i,j}\bigr)^2\,\chi_i(x)\,dx.
\label{eq:mc_loc}
\end{align}
Altogether, the vector-valued $M$-phase LCV functional reads
\[
\min_{\Phi,\vec{\mathbf{c}},\vec{\mathbf{d}}}
\;
\lambda \mathcal{E}_{\mathrm{fid}}^{M,\mathrm{vec}}(\mathbf{H}(\Phi),\vec{\mathbf{c}})
+
\beta \mathcal{E}_{\mathrm{loc}}^{M,\mathrm{vec}}(\mathbf{H}(\Phi),\vec{\mathbf{d}})
+
\mathcal{E}_{\mathrm{per}}^m(\mathbf{H}(\Phi)),
\]
with diffuse-interface approximation
\[
\min_{\mathbf{u},\vec{\mathbf{c}},\vec{\mathbf{d}}}
\;
\lambda \mathcal{E}_{\mathrm{fid}}^{M,\mathrm{vec}}(\mathbf{u},\vec{\mathbf{c}})
+
\beta \mathcal{E}_{\mathrm{loc}}^{M,\mathrm{vec}}(\mathbf{u},\vec{\mathbf{d}})
+
\mathcal{E}_{\mathrm{GL}}^{m,\varepsilon}(\mathbf{u}).
\]

For clarity, we detail the four-phase case. The updates for the region parameters are computed as
\begin{align*}
c_{1,j} &= \mathcal{A}[u_1,u_2,f_j], &
d_{1,j} &= \mathcal{A}[u_1,u_2,g_k\ast f_j-f_j],\\
c_{2,j} &= \mathcal{A}[1-u_1,u_2,f_j], &
d_{2,j} &= \mathcal{A}[1-u_1,u_2,g_k\ast f_j-f_j],\\
c_{3,j} &= \mathcal{A}[u_1,1-u_2,f_j], &
d_{3,j} &= \mathcal{A}[u_1,1-u_2,g_k\ast f_j-f_j],\\
c_{4,j} &= \mathcal{A}[1-u_1,1-u_2,f_j], &
d_{4,j} &= \mathcal{A}[1-u_1,1-u_2,g_k\ast f_j-f_j]
\end{align*}
for each channel $j=1,\ldots,C$.
 By calculus of variations, the Euler-Lagrange equations for $u_1$ and $u_2$ are 
\begin{align}
\partial_t u_1
&=
\varepsilon \Delta u_1
-
\frac{2}{\varepsilon}W'(u_1)
-
\lambda \mathcal{L}^{\mathrm{vec}}(\vec{\mathbf{c}},u_2;\mathbf{f})
-
\beta \mathcal{L}^{\mathrm{vec}}(\vec{\mathbf{d}},u_2;g_k\ast\mathbf{f}-\mathbf{f}),
\\
\partial_t u_2
&=
\varepsilon \Delta u_2
-
\frac{2}{\varepsilon}W'(u_2)
-
\lambda \mathcal{L}^{\mathrm{vec}}(\vec{\bar{\mathbf{c}}},u_1;\mathbf{f})
-
\beta \mathcal{L}^{\mathrm{vec}}(\vec{\bar{\mathbf{d}}},u_1;g_k\ast\mathbf{f}-\mathbf{f}),
\end{align}
where
\begin{align}
\mathcal{L}^{\mathrm{vec}}(\vec{\mathbf{c}},u;\mathbf{f})
&=
\|\mathbf{f}-\mathbf{c}_1\|_2^2\,u
-
\|\mathbf{f}-\mathbf{c}_2\|_2^2\,u
+
\|\mathbf{f}-\mathbf{c}_3\|_2^2\,(1-u)
-
\|\mathbf{f}-\mathbf{c}_4\|_2^2\,(1-u),
\end{align}
and
$\vec{\bar{\mathbf{c}}}=(\mathbf{c}_1,\mathbf{c}_3,\mathbf{c}_2,\mathbf{c}_4)$
and $\vec{\bar{\mathbf{d}}}=(\mathbf{d}_1,\mathbf{d}_3,\mathbf{d}_2,\mathbf{d}_4)$. As in the scalar-valued case, the system is solved using the same
MBO splitting strategy with the same initialization of the phase-field variables.

\section{Numerical Examples} \label{sec:experiments}

We present numerical results obtained by applying
Algorithms~\ref{alg:twophase}--\ref{alg:fourphase} to a variety of grayscale and color images. We begin by discussing the choice of parameters, followed by a comparison between the proposed MBO-based scheme and a standard finite-difference (FD) approach for solving \eqref{eq:LCV}. We then illustrate the performance of the LCV model on several grayscale images followed by color images.

Segmentation results produced by the classical CV model ($\beta=0$) are compared with those obtained by the LCV model ($\beta>0$). Throughout all experiments, the convolution kernel $g_k$ is chosen as a
Gaussian filter with standard deviation $\sigma=5$ and window size $k=21$. In practice, this corresponds to the following implementation in Python using
the \texttt{scipy.ndimage} package:
\[
g_k \ast f
=
\texttt{gaussian\_filter}(f,\ \texttt{sigma}=5,\ \texttt{mode=`nearest'},\ \texttt{radius}=10).
\]
Throughout this section, the local image difference is defined as
$g_k\ast f - f$. The segmentation algorithms are ran up to 200 iterations, with a stopping
criterion based on a tolerance $\texttt{tol}=10^{-5}$.
Finally, to facilitate parameter selection and ensure consistency across
experiments, pixel intensities of all images are rescaled to the
interval $[0,1]$.

\begin{figure}[hb!]
\centering

\subfloat[Blood vessel (original)]{
  \includegraphics[width=1.5in]{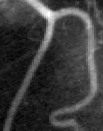}
  \label{fig:vessel}
}
\hfil
\subfloat[Topography (original)]{
  \includegraphics[width=2.5in]{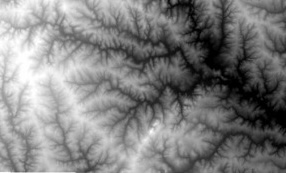}
  \label{fig:map}
}
\hfil
\subfloat[Manuscript (original)]{
  \includegraphics[width=1.5in]{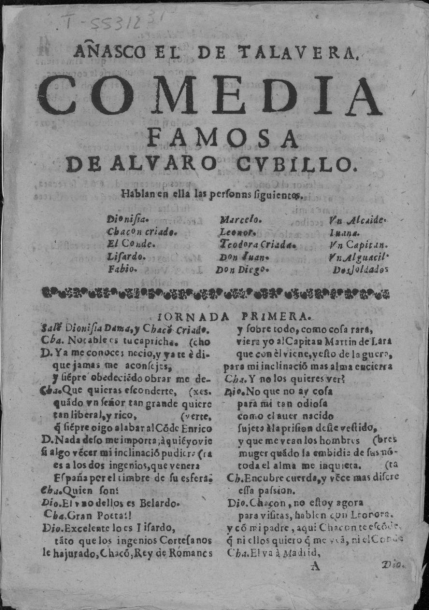}
  \label{fig:manuscript}
}

\subfloat[Blood vessel (noisy)]{
  \includegraphics[width=1.5in]{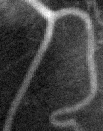}
  \label{fig:noisy_vessel}
}
\hfil
\subfloat[Topography (noisy)]{
  \includegraphics[width=2.5in]{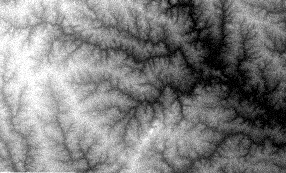}
  \label{fig:noisy_map}
}
\hfil
\subfloat[Manuscript (noisy)]{
  \includegraphics[width=1.5in]{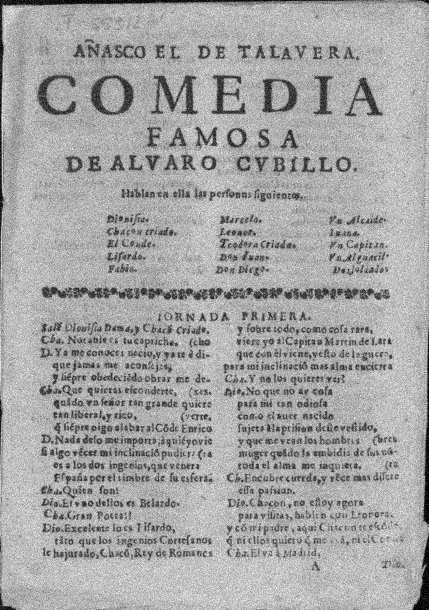}
  \label{fig:noisy_manuscript}
}

\caption{Test images for parameter analysis. 
Top row: original images whose pixel intensities are rescaled to $[0,1]$. 
Bottom row: corresponding images corrupted by Gaussian noise. 
The noisy blood vessel image has mean $0$ and standard deviation $0.01$, while the other noisy images have mean $0$ and standard deviation $0.05$.}
\label{fig:examples}
\end{figure}

\subsection{Parameter Choices} \label{sec:parameters}

For Algorithm~\ref{alg:twophase}, we first investigate the influence of the model parameters $\lambda$ and $\beta$, which balance data fidelity, regularization, and localization. Afterward, we analyze the time-step parameter $dt$, which governs stability and convergence.

To assess the robustness of these parameter choices across different domain applications, we consider three representative images: a blood vessel image, a manuscript image and a topographic map, together with their noisy versions. For the noisy versions, the blood vessel image is corrupted by additive Gaussian noise with zero mean and standard deviation $0.01$ while the topography and manuscript images are
corrupted by Gaussian noise with zero mean and standard deviation $0.05$.
All six images are shown in Figure~\ref{fig:examples}.

\subsubsection{Model Parameters $\lambda$ and $\beta$}

\begin{itemize}
    \item The \emph{fidelity parameter} $\lambda>0$ determines how closely the segmentation result approximates the original image $f$: the larger the value of $\lambda$, the more similar the segmentation is to the input image $f$.  
    
    Figure~\ref{fig:lambda-vessel} illustrates the effect of varying $\lambda$ when using Algorithm~\ref{alg:twophase} with fixed parameters $\beta=500$ and $dt=20$ on the segmentation of a blood vessel image and its noisy counterpart.
    Since $\beta>0$, the LCV model incorporates the image difference as complementary information to the original image, enhancing the contrast between the vessel and the background in both noiseless and noisy settings. This additional information is particularly effective in regions affected by intensity inhomogeneities.
    For all tested values of $\lambda$, the algorithm successfully identifies the main vessel structure, although small spurious artifacts may appear. As $\lambda$ increases, the influence of the local term progressively weakens in favor of a closer approximation to the original image. Consequently, thin or low-contrast vessel branches tend to be less well segmented. These observations suggest that intermediate values of $\lambda$ provide the best compromise between robustness to intensity inhomogeneity and preservation of fine structures.
    Moreover, the results obtained from the noisy image remain qualitatively consistent with those of the original image, highlighting the robustness of the LCV model with respect to Gaussian noise.\smallskip 
    
    \begin{figure}[h!]
     \centering
        \subfloat[Original]{\includegraphics[width=3.55cm]{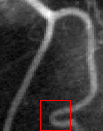}}
        \subfloat[Image difference of original]{\includegraphics[width=3.55cm]{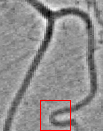}}
        \subfloat[$\lambda = 1$]{\includegraphics[width=3.55cm]{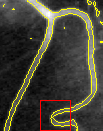}}
        \subfloat[$\lambda  =50$]{\includegraphics[width=3.55cm]{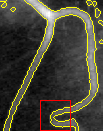}}
        \subfloat[$\lambda = 200$]{\includegraphics[width=3.55cm]{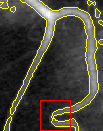}}\\
        \subfloat[Original - zoom]{\includegraphics[width=3.55cm]{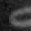}}
        \subfloat[Image difference of original - zoom]{\includegraphics[width=3.55cm]{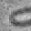}}
        \subfloat[$\lambda = 1$]{\includegraphics[width=3.55cm]{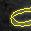}}
        \subfloat[$\lambda = 50$]{\includegraphics[width=3.55cm]{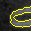}}
        \subfloat[$\lambda = 200$]{\includegraphics[width=3.55cm]{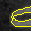}}\\
        \subfloat[Noisy]{\includegraphics[width=3.55cm]{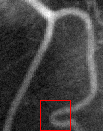}}
        \subfloat[Image difference of noisy]{\includegraphics[width=3.55cm]{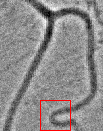}}
        \subfloat[$\lambda = 1$]{\includegraphics[width=3.55cm]{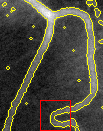}}
        \subfloat[$\lambda  =50$]{\includegraphics[width=3.55cm]{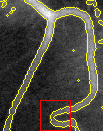}}
        \subfloat[$\lambda = 200$]{\includegraphics[width=3.55cm]{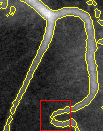}}\\
        \subfloat[Noisy - woom ]{\includegraphics[width=3.55cm]{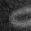}}
        \subfloat[Image difference of noisy - zoom]{\includegraphics[width=3.55cm]{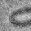}}
        \subfloat[$\lambda = 1$]{\includegraphics[width=3.55cm]{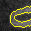}}
        \subfloat[$\lambda = 50$]{\includegraphics[width=3.55cm]{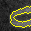}}
        \subfloat[$\lambda = 200$]{\includegraphics[width=3.55cm]{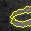}}
    \caption{Effect of varying the fidelity parameter $\lambda$ on two-phase segmentation results for a blood vessel image. Smaller values of $\lambda$ favor smoother segmentation results driven by local information, while larger values enforce stronger fidelity to the input image and may lead to partial loss of thin structures. The other parameters are fixed at  $\beta = 500$ and $dt = 20$.}
    \label{fig:lambda-vessel}
    \end{figure}
    Figure~\ref{fig:lambda-manuscript} shows the results for a manuscript image and its noisy counterpart, where the goal is to extract text information from a degraded background. This is particularly beneficial for historical documents affected by uneven illumination.
    The numerical experiments are performed with $\beta=20000$ and $dt=15$. In this example, the image difference enhances text strokes and fine character details that are weakly contrasted in the original image. 
    For all values of $\lambda$, the main text regions are mostly identified. Smaller values favor the recovery of fine textual details but may introduce spurious background components, while larger values lead to cleaner text extraction at the cost of progressively losing thin or faint characters.
    Once again, the segmentation results remain qualitatively stable under Gaussian noise, illustrating the robustness of the LCV model in document image analysis.\smallskip 
    
    \begin{figure}[h!]
     \centering
        \subfloat[Original]{\includegraphics[width=3.55cm]{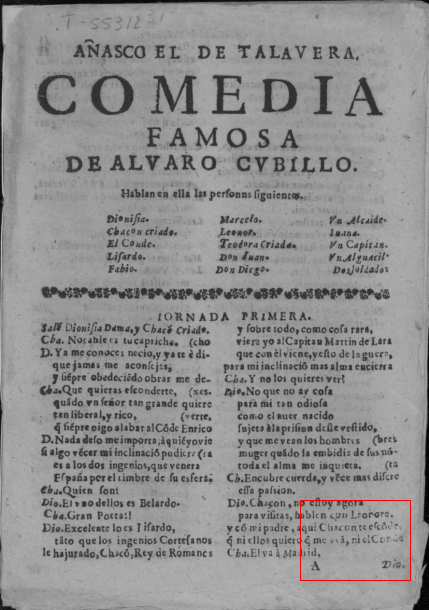}}
        \subfloat[Image difference of original]{\includegraphics[width=3.55cm]{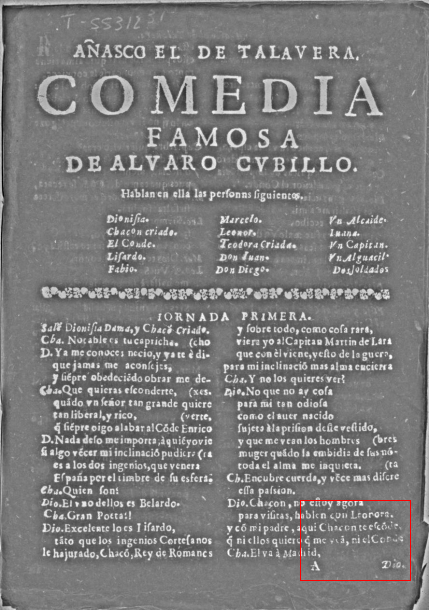}}
        \subfloat[$\lambda = 500$]{\includegraphics[width=3.55cm]{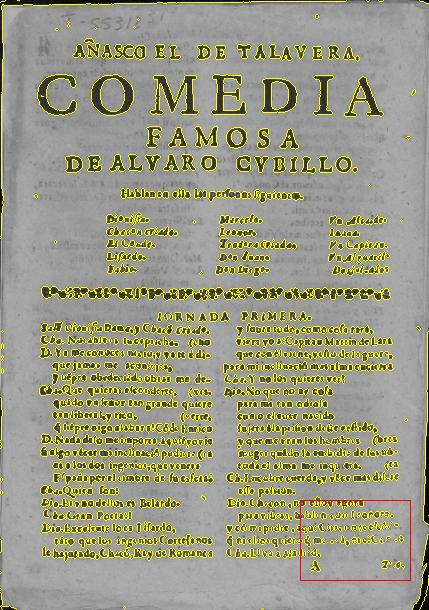}}
        \subfloat[$\lambda  =2000$]{\includegraphics[width=3.55cm]{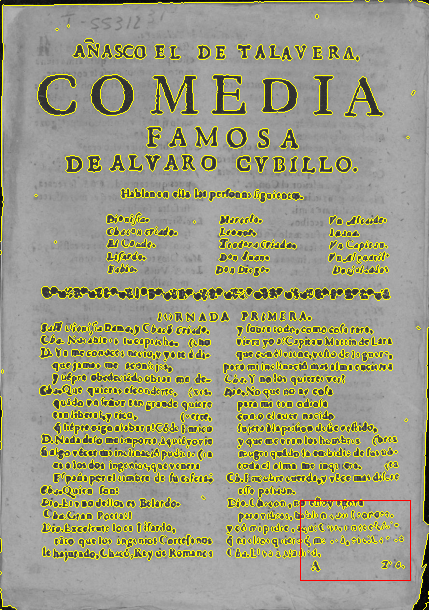}}
        \subfloat[$\lambda = 5000$]{\includegraphics[width=3.55cm]{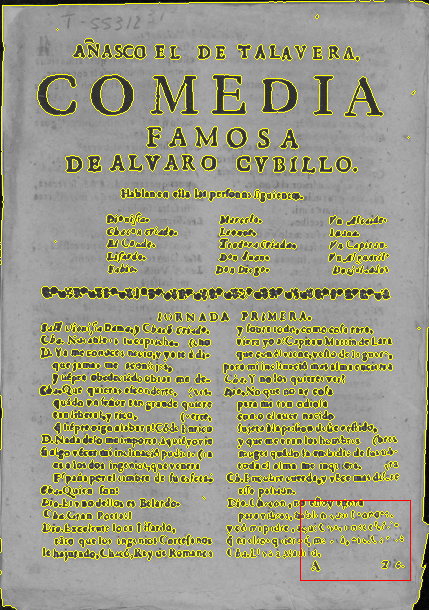}}\\
        \subfloat[Original - zoom]{\includegraphics[width=3.55cm]{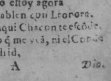}}
        \subfloat[Image difference of original - zoom]{\includegraphics[width=3.55cm]{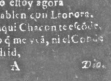}}
        \subfloat[$\lambda = 500$]{\includegraphics[width=3.55cm]{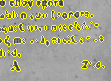}}
        \subfloat[$\lambda = 2000$]{\includegraphics[width=3.55cm]{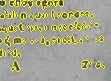}}
        \subfloat[$\lambda = 5000$]{\includegraphics[width=3.55cm]{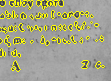}}\\
        \subfloat[Noisy]{\includegraphics[width=3.55cm]{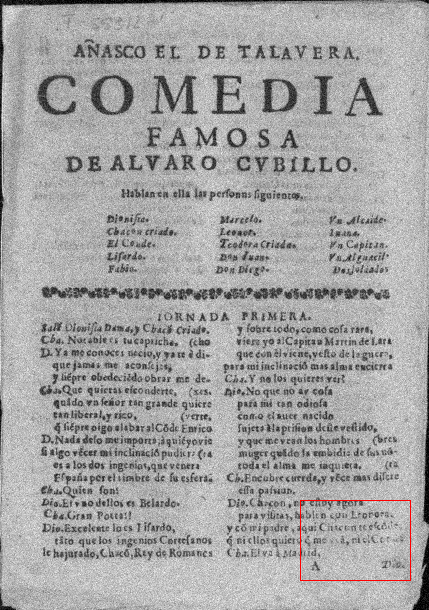}}
        \subfloat[Image difference of noisy]{\includegraphics[width=3.55cm]{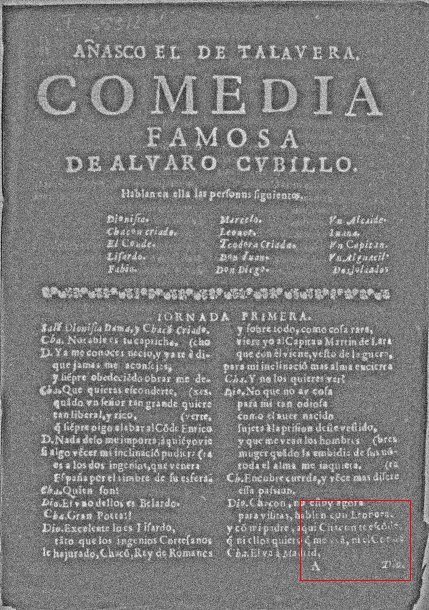}}
        \subfloat[$\lambda = 500$]{\includegraphics[width=3.55cm]{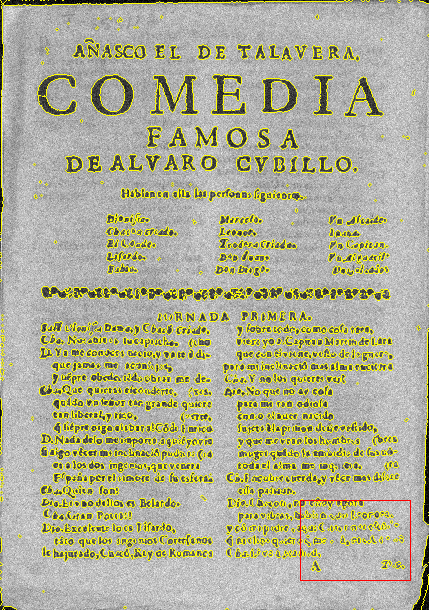}}
        \subfloat[$\lambda  =2000$]{\includegraphics[width=3.55cm]{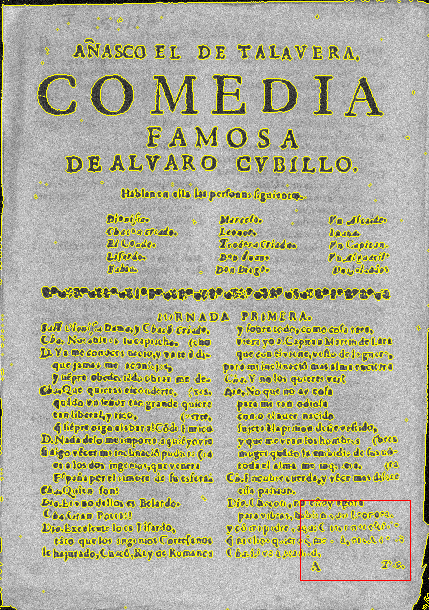}}
        \subfloat[$\lambda = 5000$]{\includegraphics[width=3.55cm]{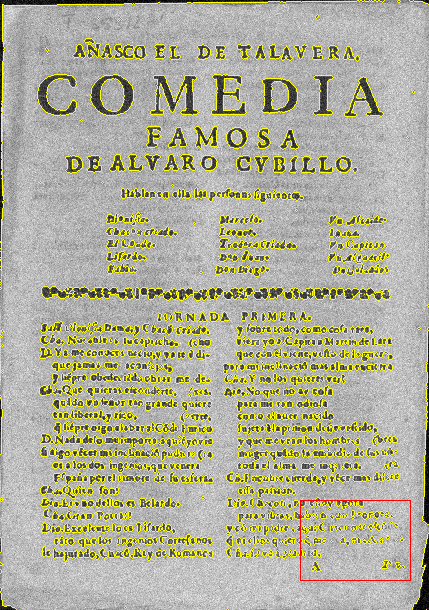}}\\
        \subfloat[Noisy - zoom]{\includegraphics[width=3.55cm]{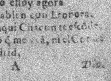}}
        \subfloat[Image difference of noisy - zoom]{\includegraphics[width=3.55cm]{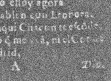}}
        \subfloat[$\lambda = 500$]{\includegraphics[width=3.55cm]{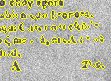}}
        \subfloat[$\lambda = 2000$]{\includegraphics[width=3.55cm]{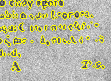}}
        \subfloat[$\lambda = 5000$]{\includegraphics[width=3.55cm]{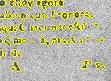}}
    \caption{Effect of varying the fidelity parameter $\lambda$ on the segmentation of a manuscript image (historical document) and its noisy counterpart. The segmentation aims at extracting text from a degraded background. Smaller values of $\lambda$ favor the detection of fine textual features but may introduce spurious components from background noise, whereas larger values lead to cleaner text extraction with a progressive loss of thin or faint characters.  The other parameters are fixed at $\beta = 20000$ and $dt = 15$.}
    \label{fig:lambda-manuscript}
    \end{figure}    
    
    Figure~\ref{fig:lambda-topo} demonstrates the influence of $\lambda$ on the segmentation of a topographic image, where gray-level intensity represents altitude (black corresponds to the lowest altitude and white to the highest). We recall that the image has been rescaled so that its gray levels take values in the normalized range $[0,1]$. The goal of the segmentation is to separate land from water, where the sharp interface corresponds to gray level $0.5$. The results are obtained with $\beta=10000$ and $dt=20$.
    In this setting, the image difference emphasizes local altitude variations and ridge--valley structures that are less apparent in the original image, thereby improving the delineation of land--water interfaces, particularly in regions where the transition is gradual or affected by illumination. The algorithm consistently captures the global land--water separation, while the level of detail in the segmented structures strongly depends on $\lambda$. Smaller values allow the segmentation to follow fine-scale terrain features, whereas larger values suppress small-scale variations and lead to coarser, more aggregated regions. As in the previous example, the segmentation results obtained from the noisy image closely resemble those of the original image across all tested values of $\lambda$, confirming the robustness of the method.
    
    \begin{figure}[t!]
     \centering
        \subfloat[Original]{\includegraphics[width=3.55cm]{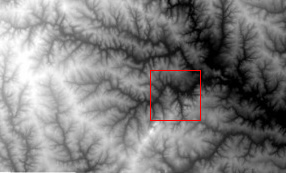}}
        \subfloat[Image difference of original]{\includegraphics[width=3.55cm]{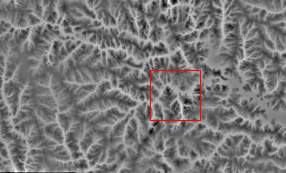}}
        \subfloat[$\lambda = 100$]{\includegraphics[width=3.55cm]{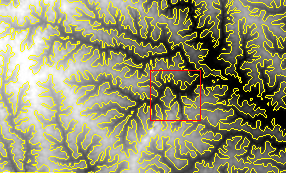}}
        \subfloat[$\lambda  =500$]{\includegraphics[width=3.55cm]{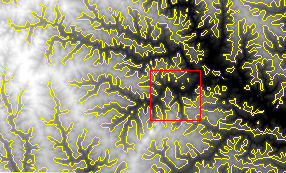}}
        \subfloat[$\lambda = 1000$]{\includegraphics[width=3.55cm]{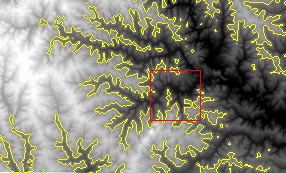}}\\
        \subfloat[Original - zoom]{\includegraphics[width=3.55cm]{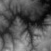}}
        \subfloat[Image difference of original - zoom]{\includegraphics[width=3.55cm]{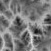}}
        \subfloat[$\lambda = 100$]{\includegraphics[width=3.55cm]{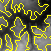}}
        \subfloat[$\lambda = 500$]{\includegraphics[width=3.55cm]{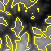}}
        \subfloat[$\lambda = 1000$]{\includegraphics[width=3.55cm]{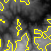}}\\
        \subfloat[Noisy]{\includegraphics[width=3.55cm]{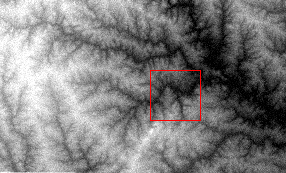}}
        \subfloat[Image difference of noisy]{\includegraphics[width=3.55cm]{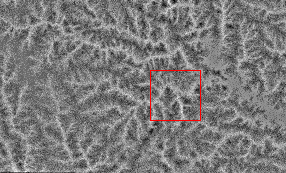}}
        \subfloat[$\lambda = 100$]{\includegraphics[width=3.55cm]{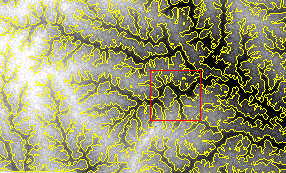}}
        \subfloat[$\lambda  =500$]{\includegraphics[width=3.55cm]{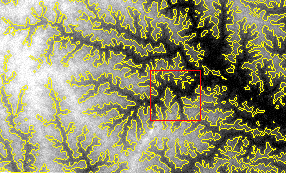}}
        \subfloat[$\lambda = 1000$]{\includegraphics[width=3.55cm]{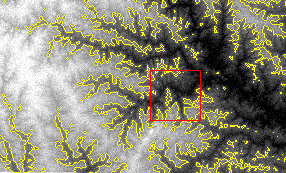}}\\
        \subfloat[Noisy - zoom ]{\includegraphics[width=3.55cm]{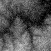}}
        \subfloat[Image difference of noisy - zoom]{\includegraphics[width=3.55cm]{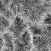}}
        \subfloat[$\lambda = 100$]{\includegraphics[width=3.55cm]{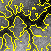}}
        \subfloat[$\lambda = 500$]{\includegraphics[width=3.55cm]{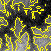}}
        \subfloat[$\lambda = 1000$]{\includegraphics[width=3.55cm]{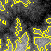}}
    \caption{Effect of varying the fidelity parameter $\lambda$ on the segmentation of a topography image, where gray-level intensity represents altitude. The task is to separate land from water. The other parameters are fixed at $\beta = 10000$ and $dt = 20$. }
    \label{fig:lambda-topo}
    \end{figure}

\newpage 
\item The \emph{localization parameter} $\beta\geq 0$ is  responsible for mitigating intensity inhomogeneity by controlling the contribution of the image-difference term in the LCV model.
    The image difference emphasizes local variations by suppressing slow spatial intensity changes, resulting in more homogeneous regions than those of the original image. Hence, it allows the LCV model to better distinguish meaningful structures from background variations caused by illumination bias. When the input image exhibits intensity inhomogeneity, it is recommended to choose $\beta > \lambda$ so that the segmentation is driven primarily by local contrast information rather than global intensity values in the original image input. Conversely, for images with relatively homogeneous intensities, one may take  $\beta \leq \lambda$ without loss of performance \cite{wang2010efficient}. 
    Overall, the stronger the intensity inhomogeneity present in the image, the larger the value $\beta$ should be used.\smallskip

    Figure~\ref{fig:beta-vessel} shows the effect of varying $\beta$ on the segmentation of the blood vessel image and its noisy counterpart when using Algorithm~\ref{alg:twophase}. The other parameters are fixed at $\lambda=50$ and $dt=20$.
    For $\beta=0$, corresponding to the classical CV model, the pronounced intensity inhomogeneity along the vessel leads to incomplete and disconnected segmentation, particularly visible in the zoomed-in views.
    As $\beta$ increases, the contribution of the image difference becomes more significant, enabling the model to better exploit local contrast information. This leads to a progressive improvement in vessel continuity and delineation. In particular, for $\beta = 500$, the blood vessel is fully segmented in both the original and noisy images, and the gaps observed for smaller values of $\beta$ are closed in the close-up views.
    This example clearly highlights the role of $\beta$ in compensating for intensity inhomogeneity. \smallskip 

    \begin{figure}[h!] 
    \centering
        \subfloat[Original]{\includegraphics[width=3.55cm]{images_python/vessel2.png}}
        \subfloat[Image difference of original]{\includegraphics[width=3.55cm]{images_python/vessel_diff.png}}
        \subfloat[$\beta = 0$]{\includegraphics[width=3.55cm]{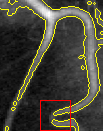}}
        \subfloat[$\beta  =100$]{\includegraphics[width=3.55cm]{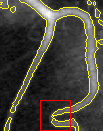}}
        \subfloat[$\beta = 500$]{\includegraphics[width=3.55cm]{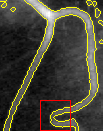}}\\
        \subfloat[Original]{\includegraphics[width=3.55cm]{images_python/close_up_vessel2.png}} 
        \subfloat[Image difference of original]{\includegraphics[width=3.55cm]{images_python/close_up_vessel_diff.png}}
        \subfloat[$\beta = 0$]{\includegraphics[width=3.55cm]{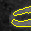}}
        \subfloat[$\beta = 100$]{\includegraphics[width=3.55cm]{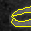}}
        \subfloat[$\beta =500$]{\includegraphics[width=3.55cm]{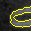}}\\
        \subfloat[Noisy]{\includegraphics[width=3.55cm]{images_python/noisy_vessel2.png}}
        \subfloat[Image difference of noisy]{\includegraphics[width=3.55cm]{images_python/noisy_vessel_diff.png}}
        \subfloat[$\beta = 0$]{\includegraphics[width=3.55cm]{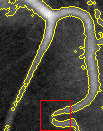}}
        \subfloat[$\beta  =100$]{\includegraphics[width=3.55cm]{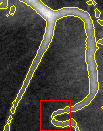}}
        \subfloat[$\beta = 500$]{\includegraphics[width=3.55cm]{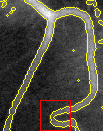}}\\
        \subfloat[Noisy]{\includegraphics[width=3.55cm]{images_python/close_up_noisy_vessel2.png}} 
        \subfloat[Image difference of noisy]{\includegraphics[width=3.55cm]{images_python/close_up_noisy_vessel_diff.png}}
        \subfloat[$\beta = 0$]{\includegraphics[width=3.55cm]{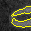}}
        \subfloat[$\beta = 100$]{\includegraphics[width=3.55cm]{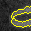}}
        \subfloat[$\beta =500$]{\includegraphics[width=3.55cm]{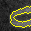}}
    \caption{Effect of varying the localization parameter $\beta$ on the segmentation of a blood vessel image and its noisy counterpart. For $\beta=0$, the classical Chan–Vese formulation yields disconnected segmentation because of intensity inhomogeneity, whereas larger values of $\beta$ progressively improve vessel continuity. The other parameters are fixed at $\lambda = 50$ and $dt = 20$.}
    \label{fig:beta-vessel}
    \end{figure}


    Figure~\ref{fig:beta-manuscript} shows the results for the manuscript image and its noisy counterpart for fixed parameter values $\lambda=2000$ and $dt=15$.
    For $\beta=0$, strong background intensity variations and illumination artifacts lead to incomplete text extraction and disconnected character strokes.
    As $\beta$ increases, the image difference enhances text strokes while suppressing slow background variations, resulting in progressively cleaner and more coherent segmentation. In particular, for large values of $\beta$, most textual components are correctly extracted and background artifacts are substantially reduced.
    As in the previous examples, increasing $\beta$ further stabilizes the segmentation in the presence of noise, yielding results that remain qualitatively consistent with those obtained from the original image.
    These observations indicate that, for document images affected by background degradation and noise, $\beta$ should be chosen  sufficiently large to achieve robust text extraction. \smallskip 

       \begin{figure}[h!] 
    \centering
        \subfloat[Original]{\includegraphics[width=3.55cm]{images_python/manuscript_image2.png}}
        \subfloat[Image difference of original]{\includegraphics[width=3.55cm]{images_python/manuscript_diff_image2.png}}
        \subfloat[$\beta = 0$]{\includegraphics[width=3.55cm]{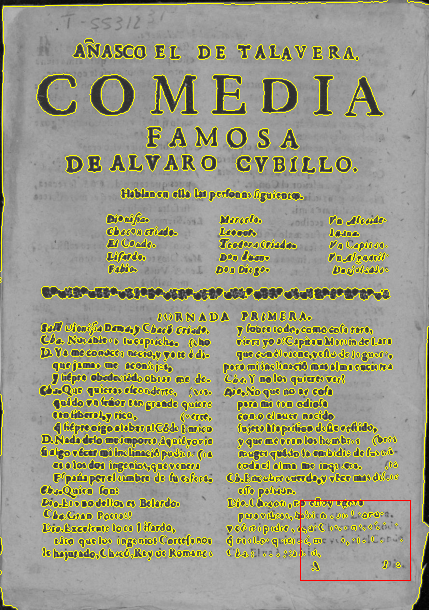}}
        \subfloat[$\beta  =2000$]{\includegraphics[width=3.55cm]{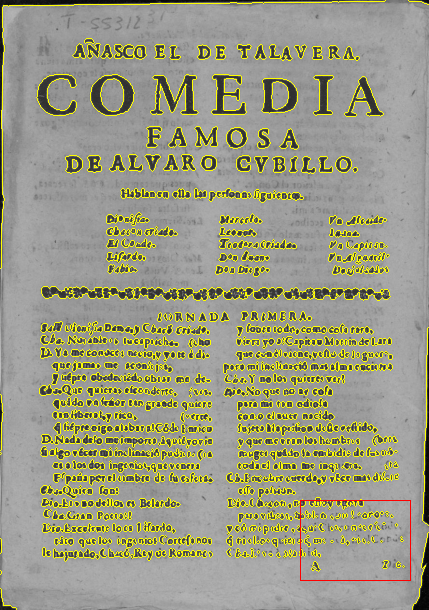}}
        \subfloat[$\beta = 20000$]{\includegraphics[width=3.55cm]{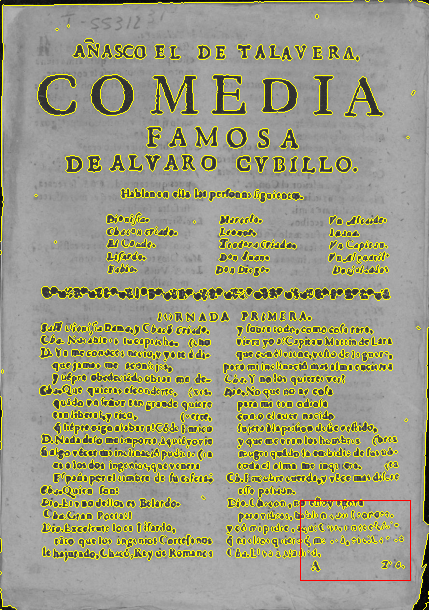}}\\
        \subfloat[Original]{\includegraphics[width=3.55cm]{images_python/close_up_manuscript_image.png}} 
        \subfloat[Image difference of original]{\includegraphics[width=3.55cm]{images_python/close_up_manuscript_diff_image.png}}
        \subfloat[$\beta = 0$]{\includegraphics[width=3.55cm]{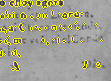}}
        \subfloat[$\beta = 2000$]{\includegraphics[width=3.55cm]{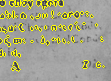}}
        \subfloat[$\beta =20000$]{\includegraphics[width=3.55cm]{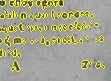}}\\
        \subfloat[Noisy]{\includegraphics[width=3.55cm]{images_python/noisy_manuscript_image2.png}}
        \subfloat[Image difference of noisy]{\includegraphics[width=3.55cm]{images_python/noisy_manuscript_diff_image2.png}}
        \subfloat[$\beta = 0$]{\includegraphics[width=3.55cm]{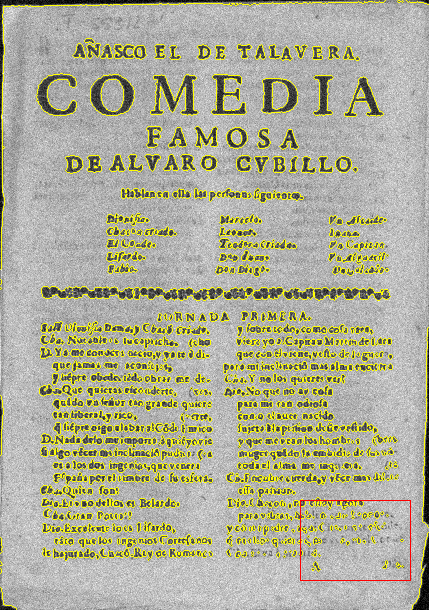}}
        \subfloat[$\beta  =2000$]{\includegraphics[width=3.55cm]{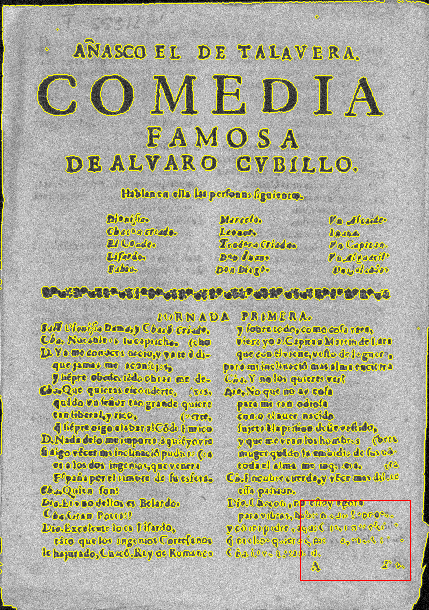}}
        \subfloat[$\beta = 20000$]{\includegraphics[width=3.55cm]{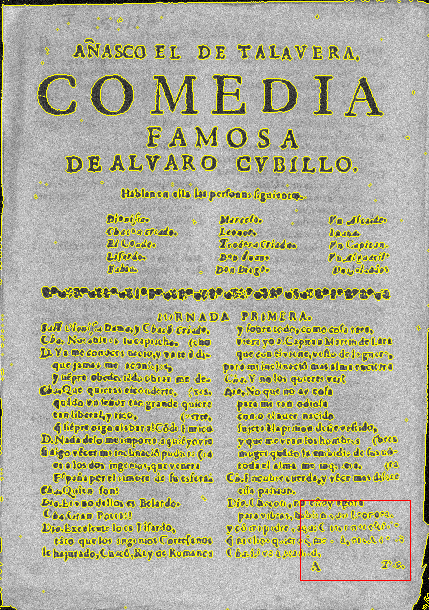}}\\
        \subfloat[Noisy]{\includegraphics[width=3.55cm]{images_python/close_up_noisy_manuscript_image.png}} 
        \subfloat[Image difference of noisy]{\includegraphics[width=3.55cm]{images_python/close_up_noisy_manuscript_diff_image.png}}
        \subfloat[$\beta = 0$]{\includegraphics[width=3.55cm]{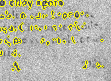}}
        \subfloat[$\beta = 2000$]{\includegraphics[width=3.55cm]{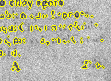}}
        \subfloat[$\beta =20000$]{\includegraphics[width=3.55cm]{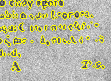}}
    \caption{Effect of varying the localization parameter $\beta$ on the segmentation of the manuscript image and its noisy counterpart. The segmentation aims at extracting textual content from a degraded background. Increasing $\beta$ enhances the contribution of the image difference, improving text extraction by compensating for background intensity inhomogeneity and illumination variations. In the presence of noise, larger values of $\beta$ further stabilize the segmentation by mitigating noise-induced intensity fluctuations. The other parameters are fixed at $\lambda = 2000$ and $dt = 15$.}
    \label{fig:beta-manuscript}
    \end{figure}

    Figure~\ref{fig:beta-topography} illustrates the influence of $\beta$ on the segmentation of the topography image and its noisy counterpart. The other parameters are set fixed to $\lambda=100$ and $dt=20$. 
    When $\beta=0$, the segmentation is strongly affected by the intensity inhomogeneity of the image, resulting in fragmented and unstable land--water interfaces.
    As $\beta$ increases, the contribution of the image difference becomes more significant, allowing the model to better exploit local altitude variations and ridge–valley structures. This results in a clearer and more coherent depiction of land–water interfaces, both in the entire images and in the zoomed-in regions. In particular, larger values of $\beta$ suppress spurious oscillations caused by slow intensity variations and improve the spatial consistency of the segmentation.
    In the presence of noise, the role of $\beta$ becomes even more pronounced. Larger values of $\beta$ further stabilize the segmentation by compensating for noise-induced intensity fluctuations. These results suggest that $\beta$ should be chosen sufficiently large for topographic data. \smallskip 
      
    \begin{figure}[t!] 
    \centering
        \subfloat[Original]{\includegraphics[width=3.55cm]{images_python/map_image2.png}}
        \subfloat[Image difference of original]{\includegraphics[width=3.55cm]{images_python/map_diff_image2.png}}
        \subfloat[$\beta = 0$]{\includegraphics[width=3.55cm]{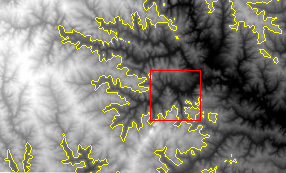}}
        \subfloat[$\beta  =1000$]{\includegraphics[width=3.55cm]{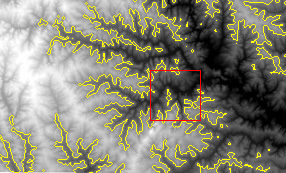}}
        \subfloat[$\beta = 10000$]{\includegraphics[width=3.55cm]{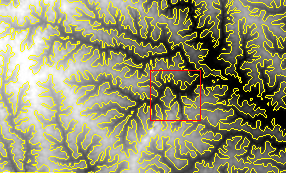}}\\
        \subfloat[Original - zoom]{\includegraphics[width=3.55cm]{images_python/close_up_map_image.png}} 
        \subfloat[Image difference of original - zoom]{\includegraphics[width=3.55cm]{images_python/close_up_map_diff_image.png}}
        \subfloat[$\beta = 0$]{\includegraphics[width=3.55cm]{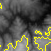}}
        \subfloat[$\beta = 1000$]{\includegraphics[width=3.55cm]{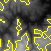}}
        \subfloat[$\beta =10000$]{\includegraphics[width=3.55cm]{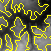}}\\
        \subfloat[Noisy]{\includegraphics[width=3.55cm]{images_python/noisy_map_image2.png}}
        \subfloat[Image difference of noisy]{\includegraphics[width=3.55cm]{images_python/noisy_map_diff_image2.png}}
        \subfloat[$\beta = 0$]{\includegraphics[width=3.55cm]{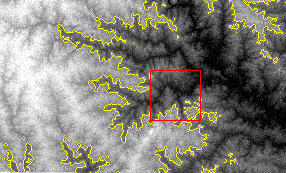}}
        \subfloat[$\beta  =1000$]{\includegraphics[width=3.55cm]{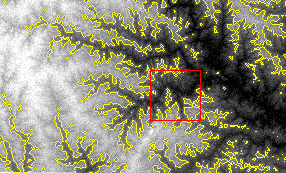}}
        \subfloat[$\beta = 10000$]{\includegraphics[width=3.55cm]{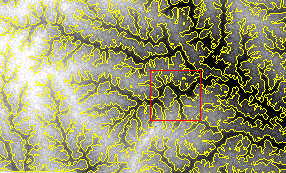}}\\
        \subfloat[Noisy - zoom]{\includegraphics[width=3.55cm]{images_python/close_up_noisy_map_image.png}} 
        \subfloat[Image difference of noisy - zoom]{\includegraphics[width=3.55cm]{images_python/close_up_noisy_map_diff_image.png}}
        \subfloat[$\beta = 0$]{\includegraphics[width=3.55cm]{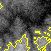}}
        \subfloat[$\beta = 1000$]{\includegraphics[width=3.55cm]{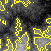}}
        \subfloat[$\beta =10000$]{\includegraphics[width=3.55cm]{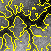}}
    \caption{Effect of varying the localization parameter $\beta$ on the segmentation of the topography image and its noisy counterpart. Increasing $\beta$ enhances the contribution of the image difference, improving the detection of land–water interfaces and stabilizing the segmentation in the presence of intensity inhomogeneity and noise. The other parameters are fixed at $\lambda = 100$ and $dt = 20$.}
    \label{fig:beta-topography}
    \end{figure}
    
\end{itemize}

\newpage 
\subsubsection{Numerical Parameter $dt$}

The \emph{numerical time step} $dt>0$ arises from the time discretization of the gradient-flow evolution associated with the Euler–Lagrange equation \eqref{eq:step1} for the LCV model. As a numerical parameter, $dt$ plays a significant role in the stability and convergence of Algorithm~\ref{alg:twophase}. \smallskip

To illustrate the impact of $dt$ in practice, we analyse below the three representative examples without any noise. We observe that large values of $dt$ may cause the evolution to overshoot, potentially leading to incomplete or inaccurate segmentation results. 
Conversely, small values of $dt$ may slow down the evolution or lead to stagnation as successive updates become too small for the segmentation to meaningfully depart from its initialization.   Thus, $dt$ should be chosen sufficiently large to allow the evolution to progress toward a stable configuration while remaining small enough to provide a reasonable approximation of the gradient flow.\smallskip
    
\begin{figure}[t!]
    \centering
        \subfloat[Original]{\includegraphics[width=3.55cm]{images_python/vessel2.png}}
        \subfloat[Image difference]{\includegraphics[width=3.55cm]{images_python/vessel_diff.png}}
        \subfloat[$\;dt = 5$]{\includegraphics[width=3.55cm]{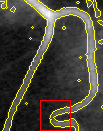}}
        \subfloat[$\;dt =20$]{\includegraphics[width=3.55cm]{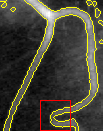}}
        \subfloat[$\;dt = 50$]{\includegraphics[width=3.55cm]{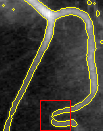}}\\
        \subfloat[Original - zoom]{\includegraphics[width=3.55cm]{images_python/close_up_vessel2.png}}
        \subfloat[Image difference - zoom]{\includegraphics[width=3.55cm]{images_python/close_up_vessel_diff.png}}
        \subfloat[$\;dt = 5$]{\includegraphics[width=3.55cm]{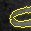}}
        \subfloat[$\;dt = 20$]{\includegraphics[width=3.55cm]{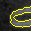}}
        \subfloat[$\;dt = 50$]{\includegraphics[width=3.55cm]{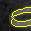}}\\
    \caption{Effect of varying the time-step size $dt$ in the MBO scheme for the LCV segmentation of the blood vessel image.  We consider $dt\in\{5,20,50\}$. For $dt=5$, spurious artifacts may appear, while for $dt=50$ the segmentation may lose fine connectivity. An intermediate choice of $dt$ yields a better balance between segmentation fidelity, numerical stability, and convergence speed. 
    The model parameters are fixed at $\lambda = 50$ and $\beta = 500$.}
    \label{fig:dt-vessel}
    \end{figure}

Figure~\ref{fig:dt-vessel} illustrates the sensitivity of Algorithm~\ref{alg:twophase}, with fixed parameters $\lambda=50$ and $\beta=500$, to the time-step parameter $dt$ for both clean and noisy blood vessel images. For $dt=5$, the evolution is too slow and may generate spurious artifacts, whereas $dt=50$ can lead to loss of connectivity. An intermediate value such as $dt=20$ provides a better balance, preserving vessel structure while limiting artifacts. \smallskip

\begin{figure}[h!]
    \centering
        \subfloat[Original]{\includegraphics[width=3.55cm]{images_python/manuscript_image2.png}}
        \subfloat[Image difference]{\includegraphics[width=3.55cm]{images_python/manuscript_diff_image2.png}}
        \subfloat[$\;dt = 10$]{\includegraphics[width=3.55cm]{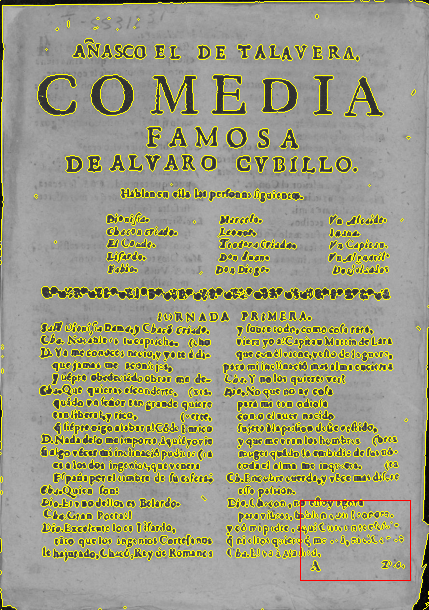}}
        \subfloat[$\;dt =15$]{\includegraphics[width=3.55cm]{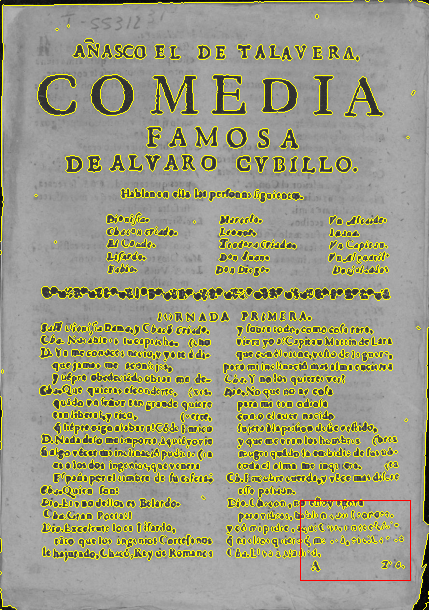}}
        \subfloat[$\;dt = 30$]{\includegraphics[width=3.55cm]{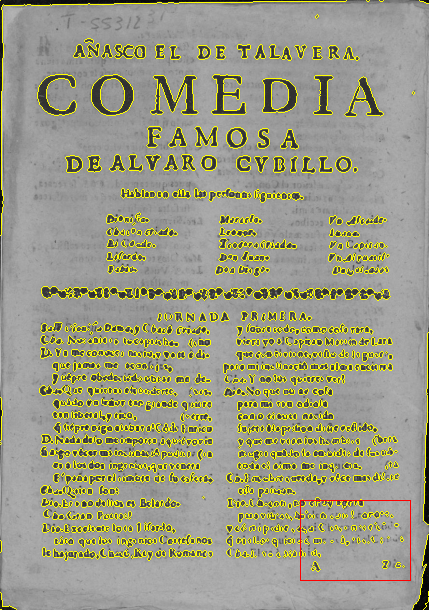}}\\
        \subfloat[Original - zoom ]{\includegraphics[width=3.55cm]{images_python/close_up_manuscript_image.png}}
        \subfloat[Image difference - zoom]{\includegraphics[width=3.55cm]{images_python/close_up_manuscript_diff_image.png}}
        \subfloat[$\;dt = 10$]{\includegraphics[width=3.55cm]{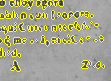}}
        \subfloat[$\;dt = 15$]{\includegraphics[width=3.55cm]{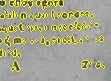}}
        \subfloat[$\;dt = 30$]{\includegraphics[width=3.55cm]{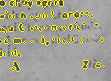}}  \\   
    \caption{Effect of varying the time-step size $dt$  in the MBO scheme for the LCV segmentation of the manuscript image.  We consider $dt \in \{10,15,30\}$.   Comparable results are obtained for $dt=10$ and $dt=15$, whereas $dt=30$ produces a coarser segmentation with fewer identified regions.  In the zoomed-in regions, larger values of $dt$ lead to the loss of finer textual details.
    The model parameters are fixed at $\lambda = 2000$ and $\beta = 20000$. }
    \label{fig:dt-text}
\end{figure}

Figure~\ref{fig:dt-text} illustrates the effect of the time-step size $dt$ on the segmentation of the manuscript image.  We consider $dt\in\{10,15,30\}$ with fixed parameter values $\lambda =2000$ and $\beta = 20000$. Comparable results are obtained for $dt=10$ and $dt=15$, whereas fewer regions are segmented for $dt=30$.  In the zoomed-in areas, less text is detected for $dt=30$ than for $dt\le 15$.  These observations indicate that excessively large time steps may induce overshooting, leading to incomplete segmentation.\smallskip

\begin{figure}[h!]
    \centering
        \subfloat[Original]{\includegraphics[width=3.55cm]{images_python/map_image2.png}}
        \subfloat[Image difference]{\includegraphics[width=3.55cm]{images_python/map_diff_image2.png}}
        \subfloat[$\;dt = 10$]{\includegraphics[width=3.55cm]{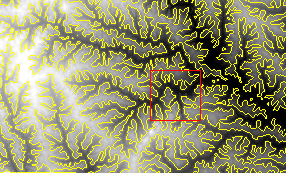}}
        \subfloat[$\;dt =20$]{\includegraphics[width=3.55cm]{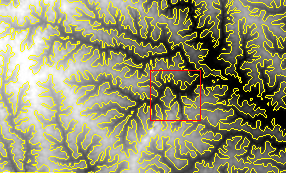}}
        \subfloat[$\;dt = 100$]{\includegraphics[width=3.55cm]{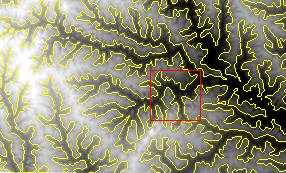}}\\
        \subfloat[Original - zoom]{\includegraphics[width=3.55cm]{images_python/close_up_map_image.png}}
        \subfloat[Image difference - zoom]{\includegraphics[width=3.55cm]{images_python/close_up_map_diff_image.png}}
        \subfloat[$\;dt = 10$]{\includegraphics[width=3.55cm]{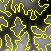}}
        \subfloat[$\;dt = 20$]{\includegraphics[width=3.55cm]{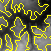}}
        \subfloat[$\;dt = 100$]{\includegraphics[width=3.55cm]{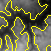}}\\
    \caption{Effect of varying the time-step size $dt$ in the MBO scheme for the LCV segmentation of the topography image. We consider $dt \in \{10,20,100\}$.  Comparable results are obtained for $dt=10$ and $dt=20$, while $dt=100$ leads to a coarser segmentation with fewer detected regions.  In the zoomed-in regions, larger values of $dt$ fail to capture finer spatial structures.
    The model parameters are fixed at $\lambda = 100$ and $\beta = 10000$. }
    \label{fig:dt-map}
    \end{figure}

Figure~\ref{fig:dt-map} examines the influence of $dt\in\{10,20,100\}$ on the segmentation of the topography image, where $\lambda = 100$ and $\beta = 10000$ are fixed. The results for $dt=10$ and $dt=20$ are comparable, while $dt=100$ yields a noticeably coarser segmentation with fewer detected features.  In the zoomed-in regions, when $dt=100$, the method captures only the main separation between land and water and fails to recover finer structures. This behavior is consistent with the overshooting effects associated with large time steps. \smallskip

To summarize our observations, Table~\ref{tab:dt_sensitivity} reports the number of iterations required to reach the stopping tolerance $\texttt{tol}=10^{-5}$ together with a qualitative assessment of the resulting segmentation behavior. All results correspond to experiments performed on the original images without noise.\smallskip

    \begin{table}[t!]
    \centering
    \begin{tabular}{lrrrcr}
    \hline
    \textbf{Example} & \textbf{$\lambda$} & \textbf{$\beta$} & \textbf{$dt$}
    & \textbf{Iterations}& \textbf{Qualitative behavior} \\
    \hline
    Blood vessel  & 50      & 500     & 5   & 13   & spurious artifacts \\
    Blood vessel  & 50      & 500     & 20  & 12   & stable \\
    Blood vessel  & 50      & 500     & 50  & 13   & disconnected regions \\
    Manuscript    & 2\,000  & 20\,000 & 10  & 12   & stable \\    
    Manuscript    & 2\,000  & 20\,000 & 15  & 15   & stable \\    
    Manuscript    & 2\,000  & 20\,000 & 30  & 21   & less text segmented \\
    Topography    & 100     & 10\,000 & 10   & 15   & spurious artifacts \\    
    Topography    & 100     & 10\,000 & 20  & 14   & stable \\
    Topography    & 100     & 10\,000 & 100 & 33   & less precise land-water interface\\
    \hline
    \end{tabular}
    \caption{Sensitivity of Algorithm~\ref{alg:twophase} to the time-step size $dt$.}
    \label{tab:dt_sensitivity}
    \end{table}

\paragraph{\bf Numerical convergence.}

\begin{figure}[h!]
    \centering
    \subfloat[Blood vessel example]{%
      \includegraphics[width=6.1cm]{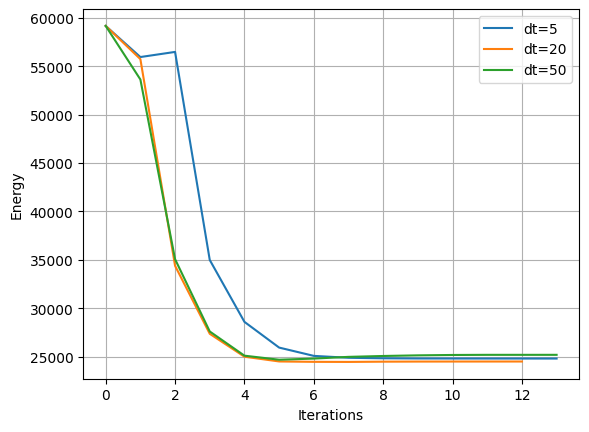}}
    \hfil
    \subfloat[Topography example]{%
      \includegraphics[width=5.9cm]{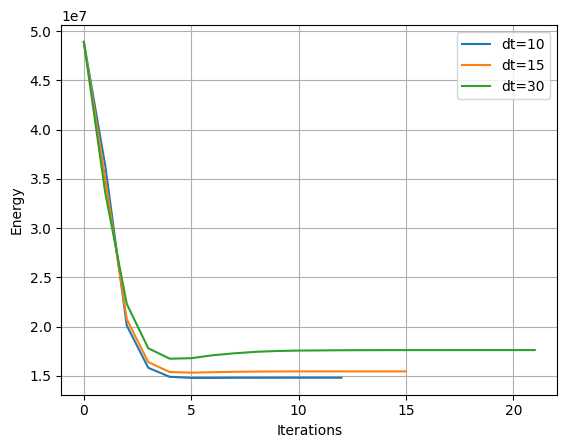}}
    \hfil
    \subfloat[Manuscript example]{%
      \includegraphics[width=5.9cm]{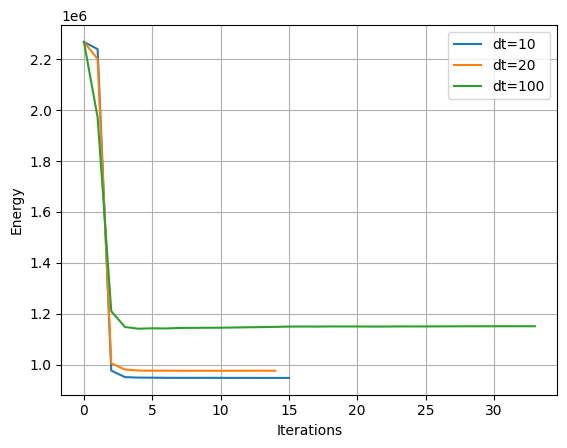}}
    \caption{The energy as a function of the iteration number for
    Algorithm~\ref{alg:twophase} for different values of $dt$ and for the three representative examples.}
    \label{fig:convergence_graph}
\end{figure}

To assess the effective numerical convergence of the scheme, we analyze the decay of the energy with respect to the iteration number for different values of $dt$.
Figure~\ref{fig:convergence_graph} shows the energy associated with \eqref{eq:LCV}, computed via Algorithm~\ref{alg:twophase}, as a function of the iterations for the same test cases as in Figures~\ref{fig:vessel}–\ref{fig:manuscript}.\smallskip

In all experiments, the energy decreases monotonically, confirming the stability of the numerical scheme. For small $dt$, the decay may be slower and sometimes may lead to stagnation. For intermediate values of $dt$, the energy decreases more rapidly and stabilizes in fewer iterations, indicating a more efficient evolution. For large $dt$, although the initial decay is fast, the algorithm may converge to a higher final energy or exhibit premature stabilization, consistent with the overshooting effects observed in the segmentation.\smallskip

Overall, these results highlight the strong influence of $dt$ on numerical convergence. Intermediate values offer the best trade-off between speed and accuracy, supporting the practical choice  $dt\geq 10$ adopted in this work.

\subsection{MBO Scheme vs.~Finite Difference for LCV Model}

We compare the proposed MBO scheme with the semi-implicit finite difference scheme implemented in the IPOL algorithm \cite{getreuer2012chan} for the LCV model. To ensure a fair comparison, we adapt the IPOL implementation to the LCV setting by incorporating the same local fidelity term so that both methods rely on comparable level-set formulations and minimize closely related variational energies.

In \cite{getreuer2012chan}, the perimeter term is approximated through a regularization of the Heaviside function, leading to a gradient-descent evolution discretized by finite differences (FD). Standard FD discretizations are known to introduce grid-induced anisotropy since the approximation of spatial derivatives depends on the orientation of the underlying grid. This can produce direction-dependent errors and geometric distortions of evolving interfaces, despite the existence of more sophisticated isotropic FD variants. 

In contrast, the proposed MBO-based scheme employs a spectral discretization of the diffusion step, which provides high accuracy for smooth solutions and avoids grid-induced directional bias. The diffusion is computed in the Fourier domain, yielding an isotropic treatment of spatial derivatives that is independent of grid orientation; combined with the thresholding step, this leads to a more geometrically consistent evolution of interfaces. \smallskip

\begin{figure}[h!]
    \centering
        \subfloat[Original]{\includegraphics[width=3.55cm]{images_python/vessel2.png}}
        \subfloat[MBO]{\includegraphics[width=3.55cm]{images_python/vessel_dt_20.png}}
        \subfloat[Finite Difference]{\includegraphics[width=3.55cm]{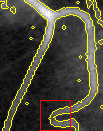}}\\
        \subfloat[Original]{\includegraphics[width=3.55cm]{images_python/close_up_vessel2.png}}
        \subfloat[MBO]{\includegraphics[width=3.55cm]{images_python/close_up_vessel_dt_20.png}}
        \subfloat[Finite Difference]{\includegraphics[width=3.55cm]{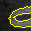}}\\
    \caption{Comparison between the MBO scheme and FD scheme on Figure \ref{fig:vessel}. Both approaches yield comparable segmentations at a global scale. The MBO scheme produces a smoother interface and suppresses small spurious artefacts, while the FD scheme preserves finer local variations.}
    \label{fig:vessel_mbo_vs_fd}
\end{figure}

\begin{figure}[h!]
    \centering
        \subfloat[Original]{\includegraphics[width=3.55cm]{images_python/map_image2.png}}
        \subfloat[MBO]{\includegraphics[width=3.55cm]{images_python/map_dt_20.png}}
        \subfloat[Finite Difference]{\includegraphics[width=3.55cm]{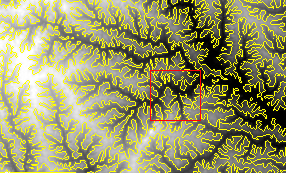}}\\
        \subfloat[Original]{\includegraphics[width=3.55cm]{images_python/close_up_map_image.png}}
        \subfloat[MBO]{\includegraphics[width=3.55cm]{images_python/close_up_map_dt_20.png}}
        \subfloat[Finite Difference]{\includegraphics[width=3.55cm]{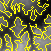}}\\
    \caption{Comparison beween MBO scheme and FD scheme on Figure \ref{fig:map}. The FD scheme yields irregular and locally fragmented interfaces along ridge–valley structures, whereas the MBO scheme produces smoother and more regular segmentations.}
    \label{fig:map_mbo_vs_fd}
\end{figure}

Including the local fidelity term in both methods the qualitative segmentation results obtained by the MBO scheme and the FD scheme are comparable, as illustrated in Figures~\ref{fig:vessel_mbo_vs_fd}-\ref{fig:map_mbo_vs_fd}. A closer inspection of these figures reveals practical differences between the two numerical approaches.\smallskip 

Figure~\ref{fig:vessel_mbo_vs_fd} shows that, for the blood vessel image, both methods successfully capture the main vessel structure and yield similar results at a global scale. The MBO-based segmentation, however, produces a slightly smoother and more regular interface and suppresses small spurious artifacts that are visible in the FD result. This behavior can be attributed to the combined effect of spectral diffusion and thresholding in the MBO scheme, which naturally regularizes fine-scale oscillations while preserving the principal geometric features of the interface.\smallskip

Figure~\ref{fig:map_mbo_vs_fd} highlights the differences between the MBO and FD schemes in the topography example. The FD scheme yields irregular and locally fragmented interfaces along ridge–valley structures, whereas the MBO scheme produces smoother and more regular boundaries at the cost of attenuating fine terrain details. These effects are particularly visible in the zoomed-in regions and reflect the respective sensitivity of the two schemes to grid-induced directional effects and diffusion strength. \smallskip 

\begin{figure}[h!]
\centering
\subfloat[Figure \ref{fig:vessel}]{  \includegraphics[width=6.7cm]{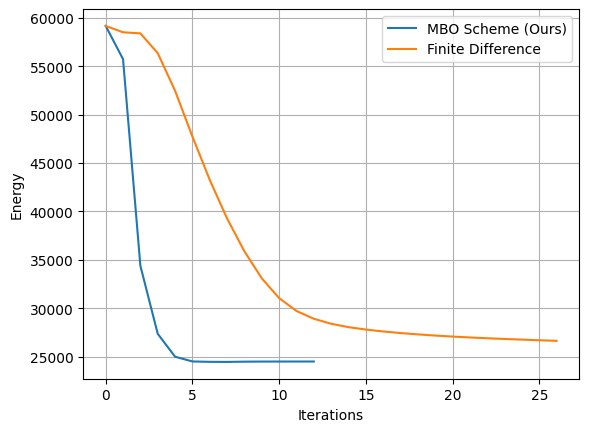}}\label{fig:vessel_fd_mbo} 
\subfloat[Figure \ref{fig:map}]{  \includegraphics[width=6.5cm]{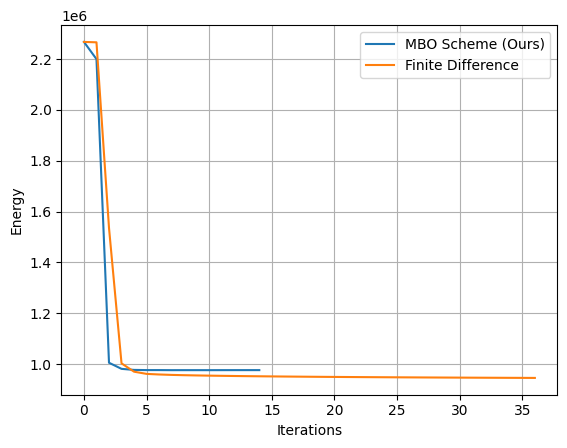}\label{fig:map_fd_mbo}} 
\caption{Numerical convergence for  MBO and FD. The energy decay is monotonic for both methods, while the MBO scheme reaches a stable configuration in fewer iterations.} \label{fig:convergence_graph2}
\end{figure}

The differences between the two approaches lie primarily in their numerical behavior rather than in the final segmented interfaces. This distinction becomes more apparent when examining numerical convergence. As shown in Figure~\ref{fig:convergence_graph2}, both schemes exhibit monotonic energy decay; however, the MBO scheme consistently reaches a stable configuration in fewer iterations than the FD scheme. This accelerated convergence of the MBO approach can be attributed to the combined effect of the thresholding mechanism and the spectral diffusion step, which rapidly enforces sharp interfaces. \smallskip

In conclusion, the main advantage of the proposed approach based on the MBO scheme lies in achieving satisfactory segmentation results with improved numerical efficiency. Hence, it serves as a robust and computationally efficient alternative to the standard FD approach for the LCV model.

\newpage\subsection{Grayscale Image Segmentation}

We begin with two-phase segmentation on a medical image to illustrate the basic behavior of the method and then turn to multiphase grayscale segmentation on STM images. The examples in this section focus on images exhibiting significant intensity inhomogeneity, a setting in which LCV models are especially well suited.

\subsubsection{Two-Phase Segmentation on MR image}

We apply Algorithm \ref{alg:twophase} to a grayscale magnetic resonance (MR) image of the brain. MR images typically suffer from intensity inhomogeneity caused by nonuniform magnetic fields and spatial variations in tissue susceptibility.

\begin{figure}[h!]
\centering
    \subfloat[Original]{\includegraphics[scale=1]{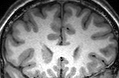}}
    \subfloat[Image difference]{\includegraphics[scale=1]{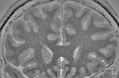}}
    \subfloat[CV, $\beta = 0$]{\includegraphics[scale=1]{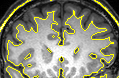}}
    \subfloat[LCV, $\beta = 40$]{\includegraphics[scale=1]{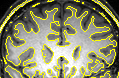}}
\caption{MRI Segmentation using the proposed MBO scheme.
Both CV and LCV methods are run with $\lambda = 5$ and $dt = 15$.
Because of its global piecewise-constant assumption, the CV model under-segments white matter regions affected by intensity inhomogeneity. On the other hand, the LCV model adapts to local intensity variations and yields a more accurate segmentation.}
\label{fig:brain}
\end{figure}

Figure~\ref{fig:brain} displays the segmentation results obtained with parameters $\lambda = 5.0$ and $dt = 15$. For the LCV model, we set $\beta = 40$. As expected, the classical CV model under-segments the white matter regions because of its global intensity assumption. In contrast, the LCV model successfully adapts to local intensity variations and yields a more accurate segmentation of anatomical structures.
This example highlights the robustness of the proposed MBO scheme when combined with local fidelity terms, particularly in the presence of strong bias fields.

\subsubsection{Four-Phase Grayscale Segmentation}

\begin{figure}[h!]	
\centering
	\subfloat[Original]{\includegraphics[width=3cm]{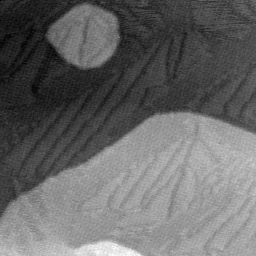}}
    \subfloat[CV, $\beta=0$]{\includegraphics[width=3cm]{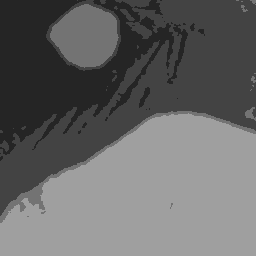}}	
    \subfloat[Phase 1, CV]{\includegraphics[width=3cm]{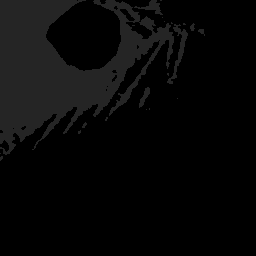}}
    \subfloat[Phase 2, CV]{\includegraphics[width=3cm]{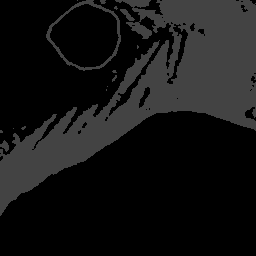}}
    \subfloat[Phase 3, CV]{\includegraphics[width=3cm]{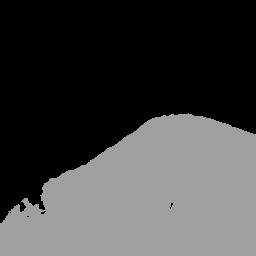}}
    \subfloat[Phase 4, CV]{\includegraphics[width=3cm]{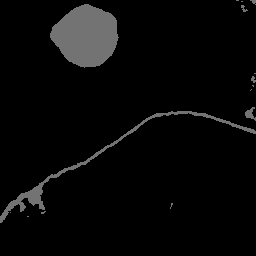}}\\  \smallskip
    \subfloat[Image difference]{\includegraphics[width=3cm]{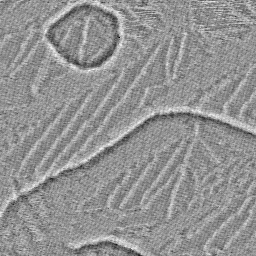}}
    \subfloat[LCV $\beta=50$]{\includegraphics[width=3cm]{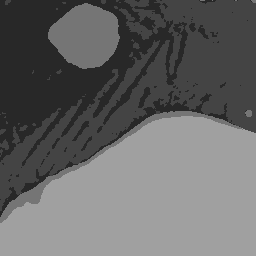}}	
    \subfloat[Phase 1, LCV]{\includegraphics[width=3cm]{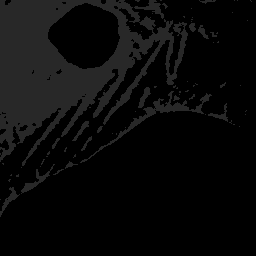}}	
    \subfloat[Phase 2, LCV]{\includegraphics[width=3cm]{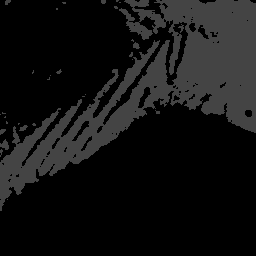}}	
    \subfloat[Phase 3, LCV]{\includegraphics[width=3cm]{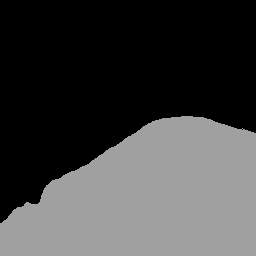}}	
    \subfloat[Phase 4, LCV]{\includegraphics[width=3cm]{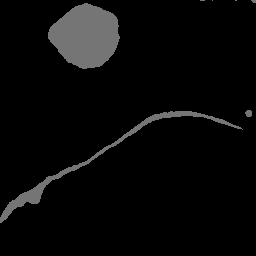}}  \smallskip
\caption{Four-phase segmentation of cyanide on Au{111} \cite{guttentag2016hexagons} along with the individual phases obtained by the CV and LCV models.
Parameters are $\lambda = 10$ and $dt = 40$ for both models and $\beta = 50$ for LCV. 
While both methods produce comparable coarse segmentation results, the LCV model significantly improves the recovery of diagonal features and better preserves fine-scale variations induced by local intensity inhomogeneity, leading to more geometrically faithful phase boundaries than CV.}
\label{fig:microscopy1}
\end{figure}

\begin{figure}[h!]
\centering
	\subfloat[Original]{\includegraphics[width=3cm]{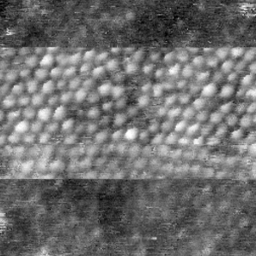}} 
    \subfloat[CV $\beta=0$]{\includegraphics[width=3cm]{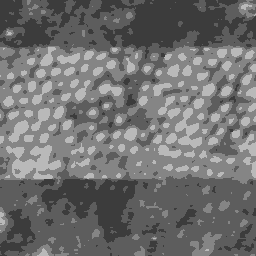}}  	
    \subfloat[Phase 1, CV]{\includegraphics[width=3cm]{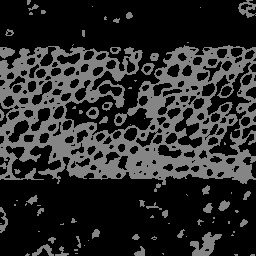}} 	
    \subfloat[Phase 2, CV]{\includegraphics[width=3cm]{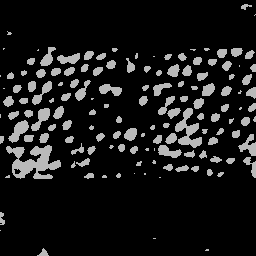}}	
    \subfloat[Phase 3, CV]{\includegraphics[width=3cm]{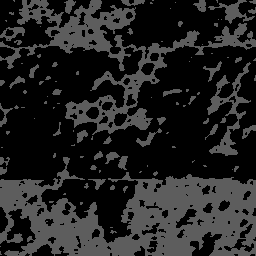}}	
    \subfloat[Phase 4, CV]{\includegraphics[width=3cm]{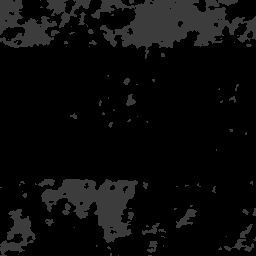}}\\  \smallskip
    \subfloat[Image difference]{\includegraphics[width=3cm]{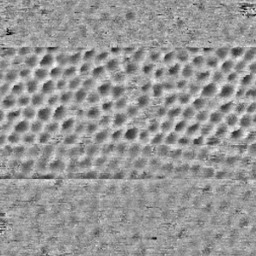}}
    \subfloat[LCV, $\beta=80$]{\includegraphics[width=3cm]{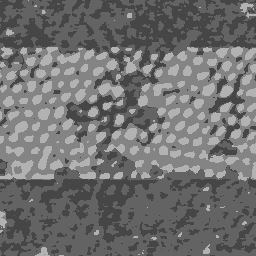}}  
    \subfloat[Phase 1, LCV]{\includegraphics[width=3cm]{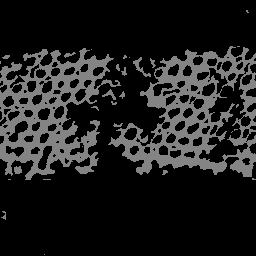}} 	
    \subfloat[Phase 2, LCV]{\includegraphics[width=3cm]{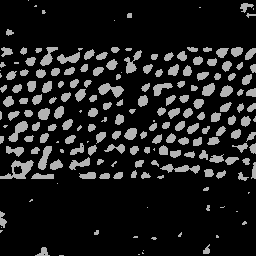}}	
    \subfloat[Phase 3, LCV]{\includegraphics[width=3cm]{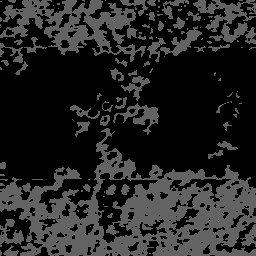}}	
    \subfloat[Phase 4, LCV]{\includegraphics[width=3cm]{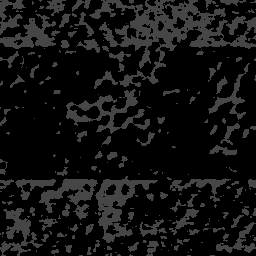}}  \smallskip
\caption{Four-phase segmentation results for 9,12-carboranedithiol on Au{111}/mica \cite{thomas2018acid} with texture and intensity variation along with the individual phases obtained by the CV and LCV models.
Parameters are $\lambda = 20$ and $dt = 30$ for both models and $\beta = 80$ for LCV. 
Compared with CV, the LCV model segments a substantially larger portion of the textured region, resolves finer structural details, and yields more coherent phase boundaries in areas affected by spatially varying intensities.}
\label{fig:microscopy2}
\end{figure}

We next demonstrate the ability of the multiphase LCV model \eqref{eq:M-LCV} to handle multiphase grayscale segmentation on two STM images.
Figures~\ref{fig:microscopy1} and~\ref{fig:microscopy2} display STM images of cyanide on Au{111} \cite{guttentag2016hexagons} and 9,12-carboranedithiol on Au{111}/mica \cite{thomas2018acid}, respectively. For both images, we compare CV and LCV segmentations using identical parameters $\lambda=10$ and $dt=40$ while introducing the local parameter $\beta=50$ for LCV. Although the resulting piecewise-constant approximations appear visually similar at first glance, closer inspection reveals clear, qualitative improvements achieved by the LCV model.
In particular, Figure~\ref{fig:microscopy1} shows that the LCV model successfully captures textured regions that are either omitted or merged by the CV model, demonstrating its superior adaptability to spatial intensity variations. In
Figure~\ref{fig:microscopy2}, the LCV model segments a substantially larger portion of the central “pebbled” structure than the CV model while maintaining clear and well-defined region boundaries. Overall, these results provide strong evidence of the effectiveness of the LCV model for multiphase problems and highlight its enhanced ability to preserve fine-scale structures in the presence of spatially varying intensities, where classical global models are insufficient.

%

\newpage \subsection{Color Image Segmentation}

We now extend the method to color image segmentation. As color images can be represented in different color spaces, the choice of representation is crucial for effective segmentation.

\subsubsection{Choice of Color Space}
Color images can be represented in many color spaces such as RGB (red, green, blue), HSV (hue, saturation, value), and Lab (luminosity, red–green, and yellow–blue). While RGB is widely used, its channels are highly correlated \cite{cai2017three}, which makes the $L^2$ fidelity term across channels in \eqref{eq:mc_fid}–\eqref{eq:mc_loc}  redundant. 
On the other hand, the channels in both HSV and Lab are significantly less correlated. Moreover, since these spaces are perceptually uniform \cite{paschos2001perceptually}, they are more sensitive to color differences than RGB. As a result, they generally outperform RGB in color image segmentation, particularly for texture analysis \cite{paschos2001perceptually}. 

\begin{figure}[h!]
    \centering
        \subfloat[RGB Image]{\includegraphics[width=3.55cm]{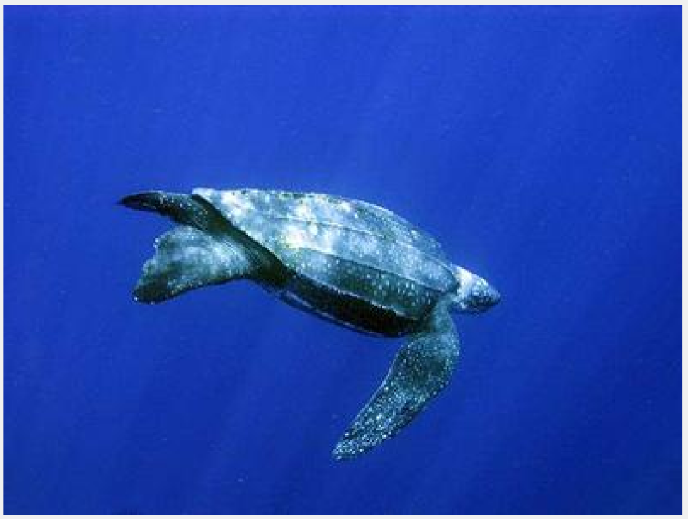}}
        \subfloat[Red Channel]{\includegraphics[width=3.55cm]{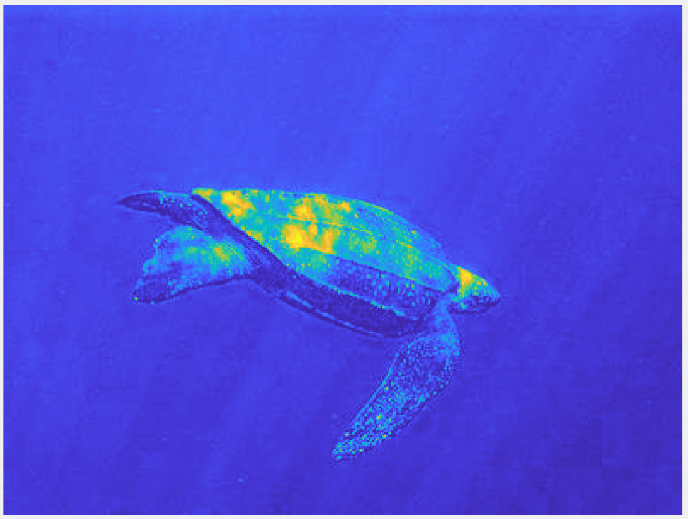}}
        \subfloat[Green Channel]{\includegraphics[width=3.55cm]{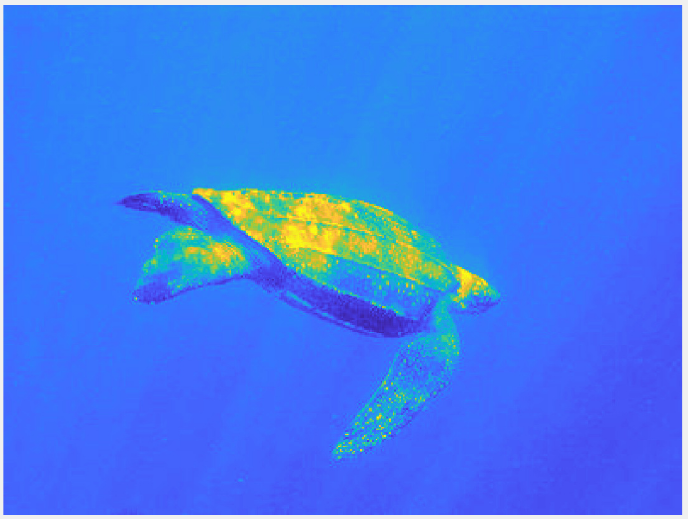}}
        \subfloat[Blue Channel]{\includegraphics[width=3.55cm]{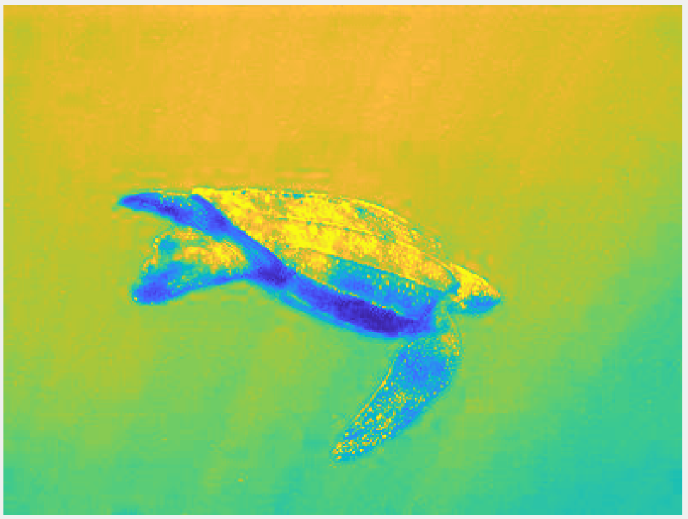}}
        \subfloat[RGB Segmentation]{\includegraphics[width=3.55cm]{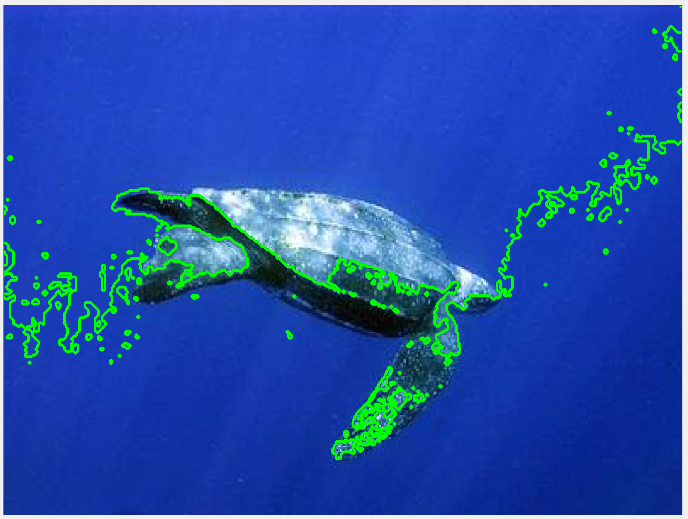}}\\
        \subfloat[HSV Image]{\includegraphics[width=3.55cm]{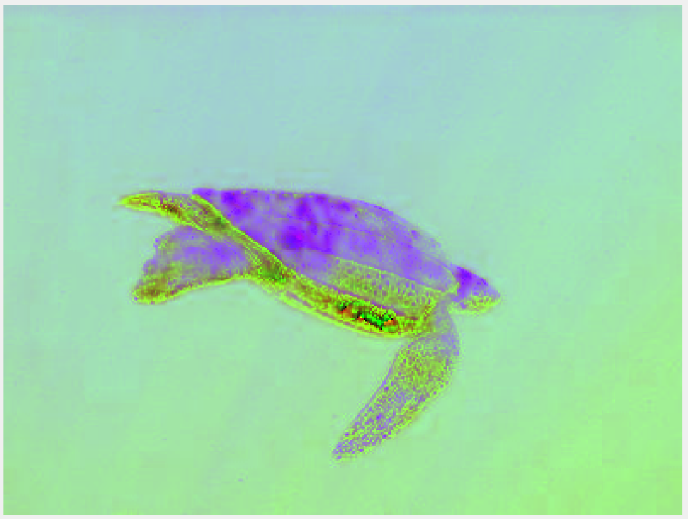}}
        \subfloat[Hue Channel]{\includegraphics[width=3.55cm]{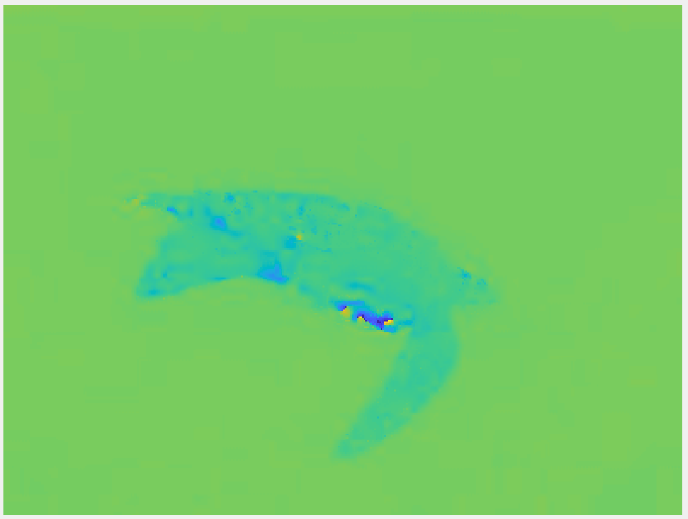}}
        \subfloat[Saturation Channel]{\includegraphics[width=3.55cm]{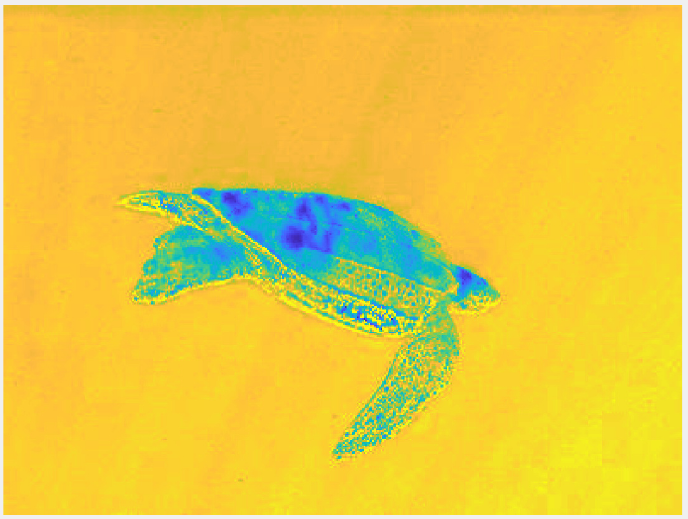}}
        \subfloat[Value Channel]{\includegraphics[width=3.55cm]{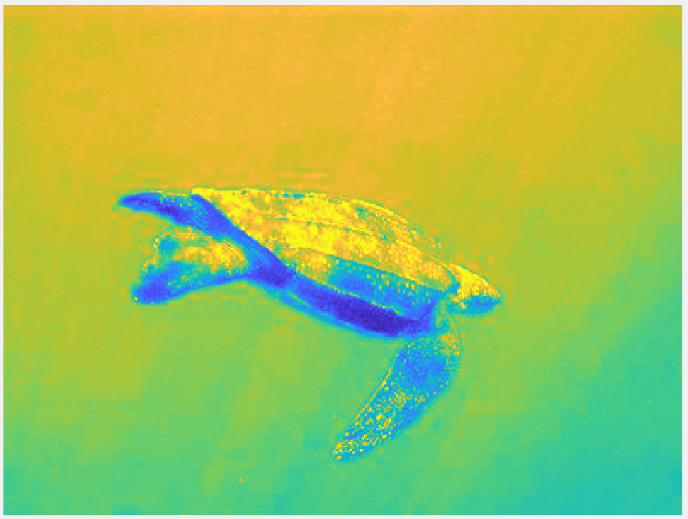}}
        \subfloat[HSV Segmentation]{\includegraphics[width=3.55cm]{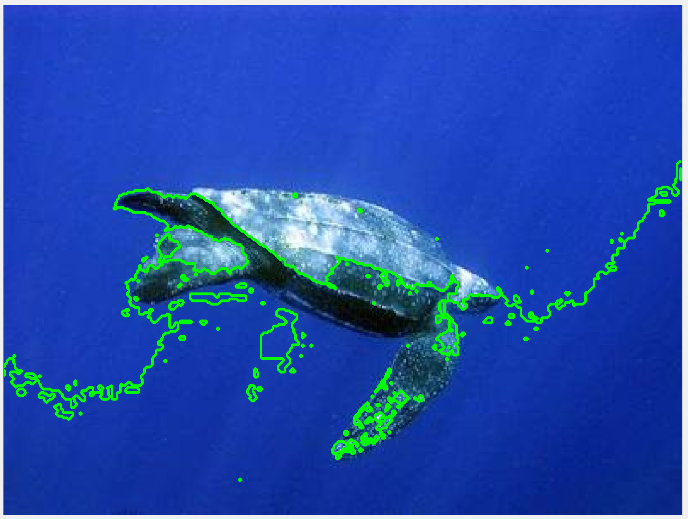}}\\
        \subfloat[Lab Image]{\includegraphics[width=3.55cm]{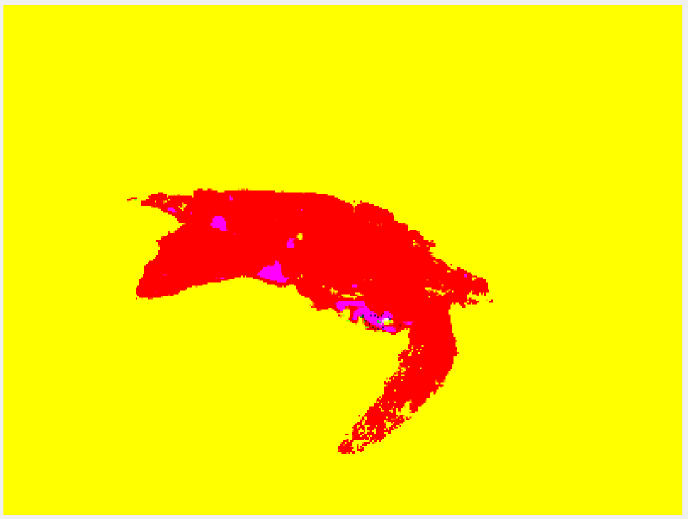}}
        \subfloat[Luminosity Channel]{\includegraphics[width=3.55cm]{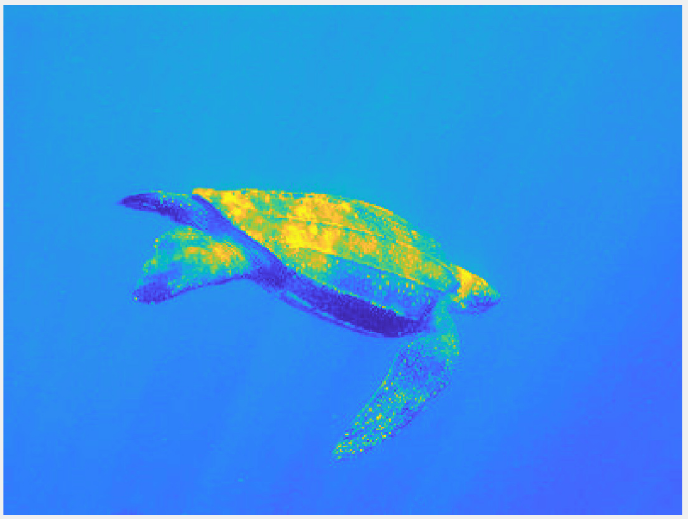}}
        \subfloat[Red-green Channel]{\includegraphics[width=3.55cm]{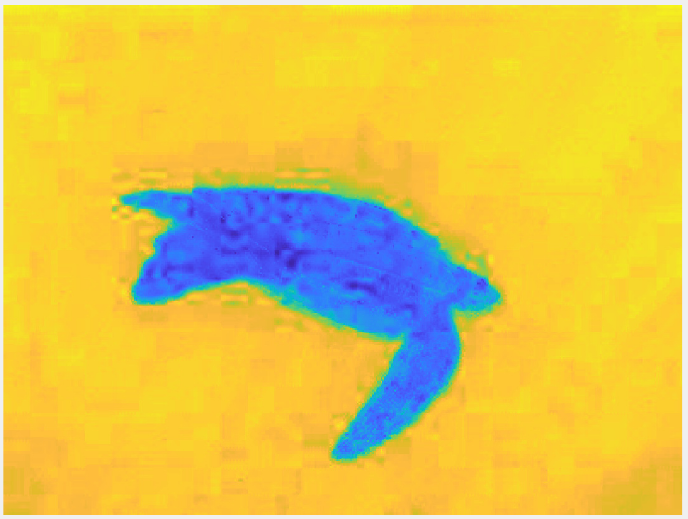}}
        \subfloat[Blue-yellow Channel]{\includegraphics[width=3.55cm]{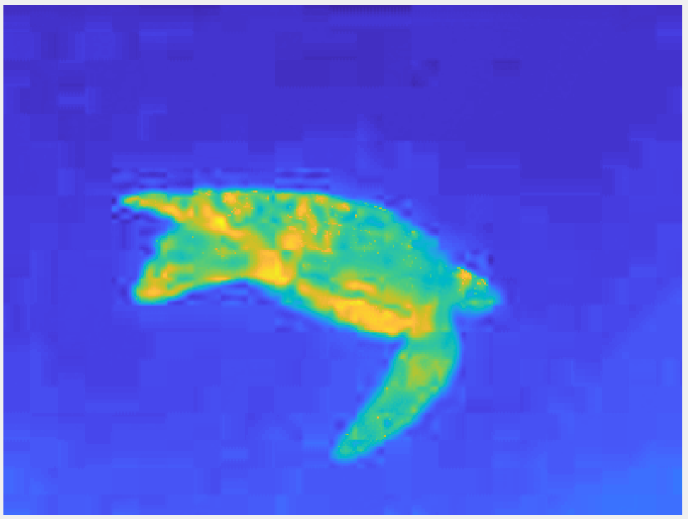}}
        \subfloat[Lab Segmentation]{\includegraphics[width=3.55cm]{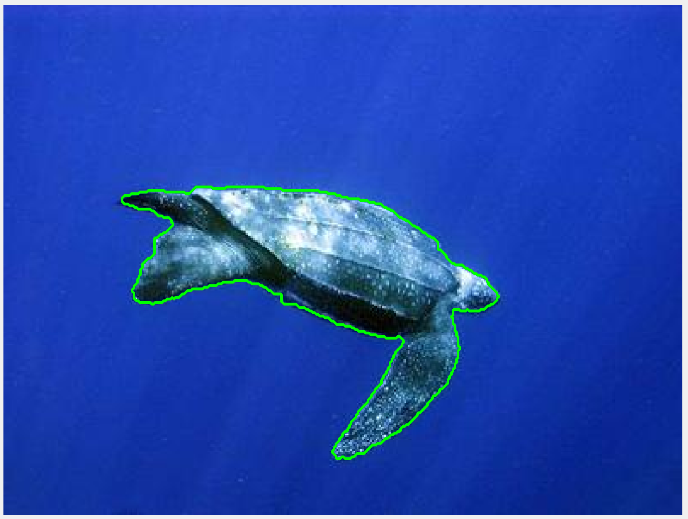}}
    \caption{Images shown in different color spaces along with their respective channels and segmentation results obtained by the CV model. Parameters are $\lambda= 30$ and $dt=20$. Because of strong inter-channel correlation, RGB segmentation is less effective. HSV improves separation but remains sensitive to illumination. Lab space yields the most consistent piecewise-constant approximation.}
\label{fig:color_space}
\end{figure}

Figure~\ref{fig:color_space} compares segmentation results obtained by applying the CV model with $\lambda=30$ and $dt=20$ in different color spaces.  Among the tested representations, the Lab color space yields the most accurate segmentation. 
This stems from the observation that, in Lab space, two channels are nearly piecewise constant, making the image structure more compatible with the CV model. The effectiveness of using Lab space for color image segmentation was examined by other works \cite{bui2024efficient, cai2017three}. 
Hence, all subsequent color experiments are performed in Lab space before applying the LCV model.

\subsubsection{Two-Phase Color Segmentation}

We apply the LCV model to natural color images exhibiting illumination and background complexity. 

\begin{figure}[h!]
\centering
	\begin{tabular}{c@{}c@{}c@{}}
		\subfloat[Original]{\includegraphics[scale=0.35]{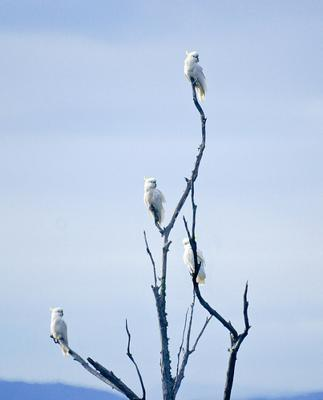}}& 
        \subfloat[CV]{\includegraphics[scale=0.35]{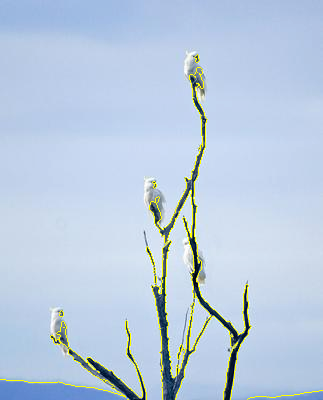}}  &	
        \subfloat[LCV]{\includegraphics[scale=0.35]{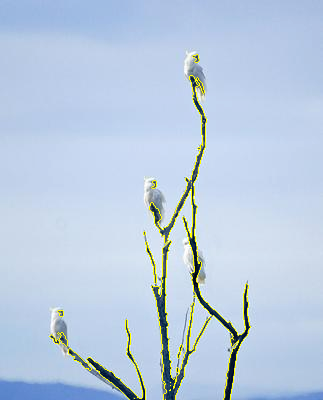}} 
	\end{tabular}
	\caption{Two-phase segmentation of a natural color image using the CV and LCV models. Parameters are $\lambda = 30$ and $\;dt = 10$ for both models and $\beta = 100$ for LCV.}
\label{fig:2phase-bird_tree}
\end{figure}

Figure \ref{fig:2phase-bird_tree} displays the  segmentation of a natural color image using the CV and LCV models. While CV merges the tree in the foreground and the hill in the background, LCV successfully isolates the tree by adapting to local color variations along   the tree branches and hill edges.

\begin{figure}[h!]
\centering
		\subfloat[Original]{\includegraphics[width=4cm]{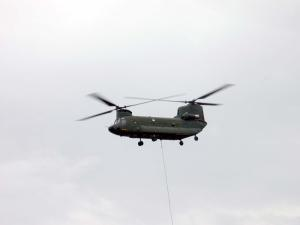}}
        \subfloat[CV]{\includegraphics[width=4cm]{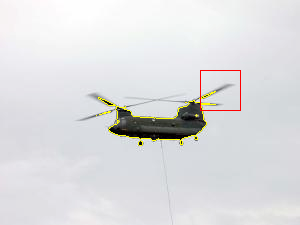}} 
        \subfloat[LCV]{\includegraphics[width=4cm]{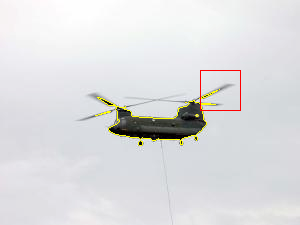}}\\
        \hspace{4cm}
        \subfloat[Close up of CV]{\includegraphics[width=4cm]{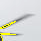}}
        \subfloat[Close up of LCV]{\includegraphics[width=4cm]{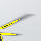}}
	\caption{Segmentation results for a color image containing thin structures and fine geometric details.
Parameters are $\lambda = 50$ and $\;dt = 20$ and $\beta = 50$ for LCV.} 
\label{fig:2phase-chopper}
\end{figure}

Figure \ref{fig:2phase-chopper} displays the segmentation results for a color image containing thin structures and fine geometric details. The LCV model better preserves thin structures such as propeller blades, whereas CV produces partial merging and loss of detail. These examples highlight the ability of the MBO-based LCV scheme to capture sharp interfaces.
        
\subsubsection{Four-Phase Color Segmentation}

\begin{figure}[h!]
\centering
\begin{tabular}{c@{\hspace{0.5cm}}c@{\hspace{0.5cm}}c}

\begin{tabular}{c}
\subfloat[Original\label{fig:church_original}]
{\includegraphics[width=5.00cm]{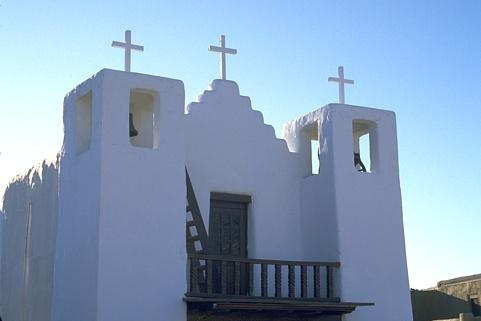}}\\
\end{tabular}
&
\begin{tabular}{c}
\subfloat[CV, $\beta=0$\label{fig:CV_approx_church}]
 {\includegraphics[width=5.00cm]{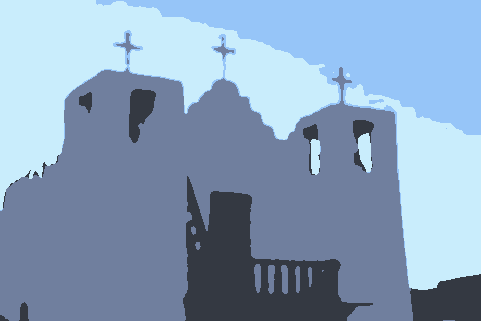}}\\[0.15cm]
\subfloat[Phase 1, CV]
 {\includegraphics[width=5.00cm]{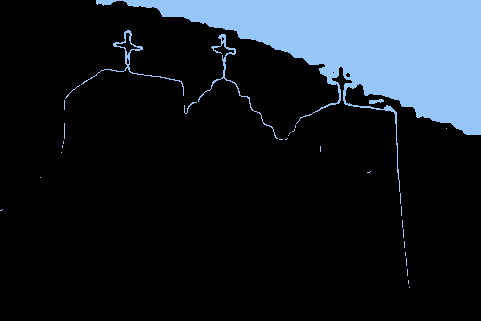}}\\
\subfloat[Phase 2, CV]
 {\includegraphics[width=5.00cm]{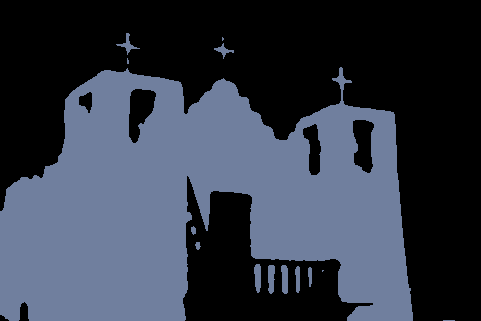}}\\
\subfloat[Phase 3, CV]
 {\includegraphics[width=5.00cm]{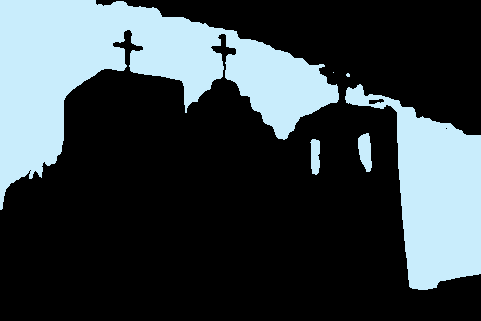}}\\
\subfloat[Phase 4, CV]
 {\includegraphics[width=5.00cm]{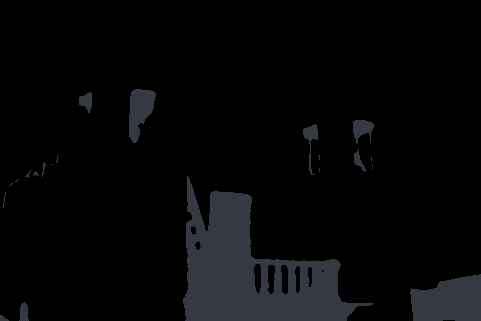}}
\end{tabular}
&
\begin{tabular}{c}
\subfloat[LCV, $\beta=200$\label{fig:LCV_approx_church}]
 {\includegraphics[width=5.00cm]{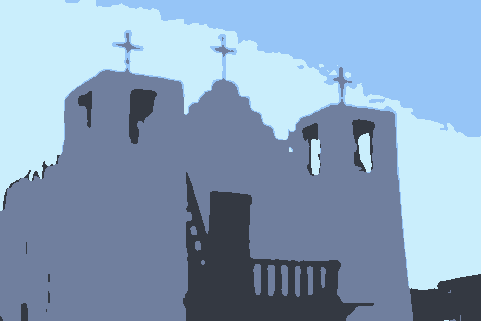}}\\[0.15cm]
\subfloat[Phase 1, LCV]
 {\includegraphics[width=5.00cm]{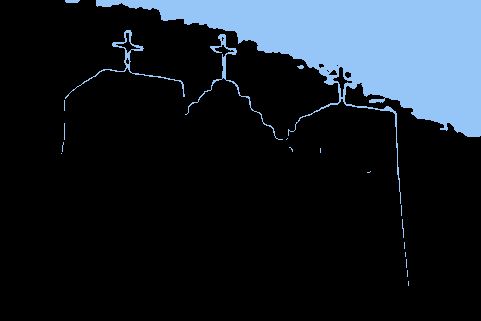}}\\
\subfloat[Phase 2, LCV]
 {\includegraphics[width=5.00cm]{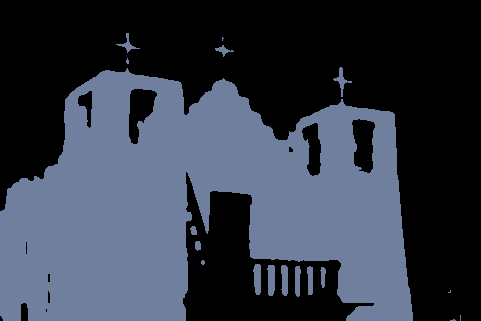}}\\
\subfloat[Phase 3, LCV]
 {\includegraphics[width=5.00cm]{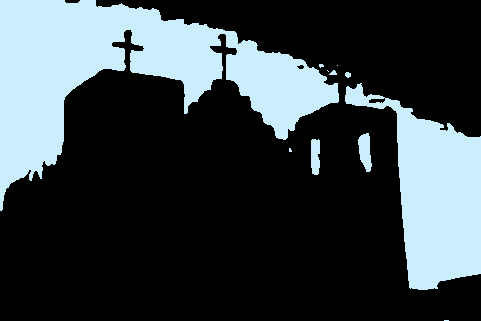}}\\
\subfloat[Phase 4, LCV]
 {\includegraphics[width=5.00cm]{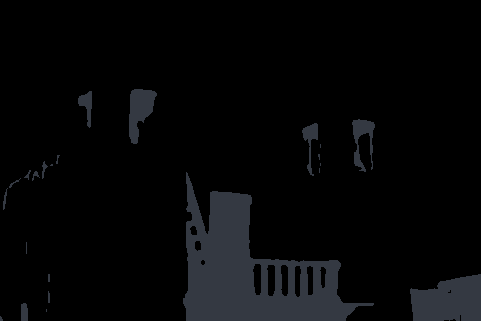}}
\end{tabular}

\end{tabular}

\caption{Four-phase segmentation of a natural color image using CV and LCV models along with the individual phases. 
Parameters are $\lambda = 40$ and $dt = 50$ for both models and $\beta = 200$ for LCV.   
While both models produce similar coarse segmentations results, LCV separates more accurately the ladder and railing structures near the entrance, avoiding artificial merging caused by illumination variations.}
\label{fig:4phase-church}
\end{figure}

\begin{figure}[h!]
\centering
\begin{tabular}{c@{\hspace{0.5cm}}c@{\hspace{0.5cm}}c}

\begin{tabular}{c}
\subfloat[Original\label{fig:orig_butterfly}]
{\includegraphics[width=5.00cm]{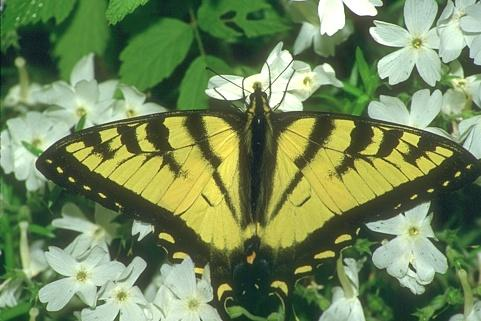}}\\
\end{tabular}
&
\begin{tabular}{c}
\subfloat[CV, $\beta=0$\label{fig:CV_approx_butterfly}]
 {\includegraphics[width=5.00cm]{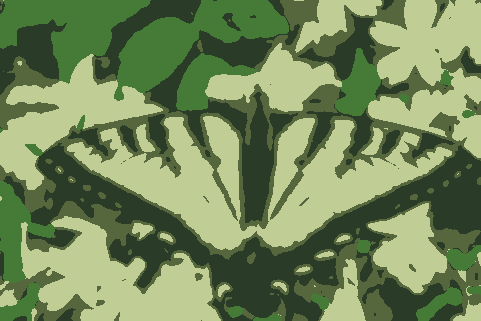}}\\[0.15cm]
\subfloat[Phase 1, CV]
 {\includegraphics[width=5.00cm]{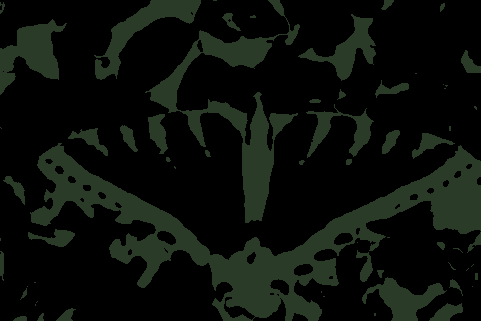}}\\
\subfloat[Phase 2, CV\label{fig:cv_phase2}]
 {\includegraphics[width=5.00cm]{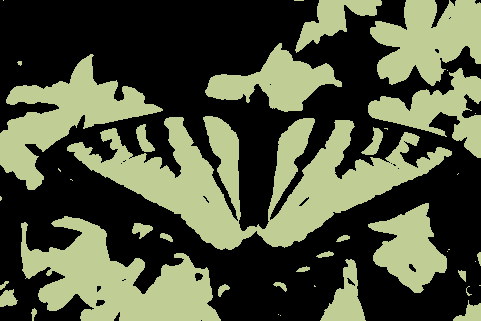}}\\
\subfloat[Phase 3, CV]
 {\includegraphics[width=5.00cm]{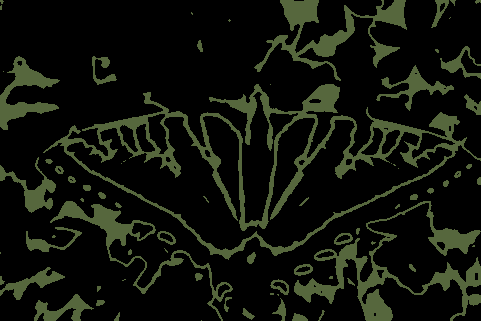}}\\
\subfloat[Phase 4, CV]
 {\includegraphics[width=5.00cm]{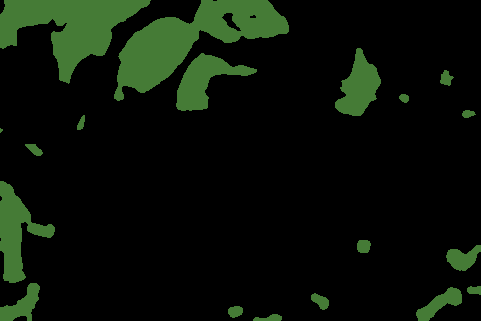}}
\end{tabular}
&
\begin{tabular}{c}
\subfloat[LCV, $\beta=60$\label{fig:LCV_approx_butterfly}]
 {\includegraphics[width=5.00cm]{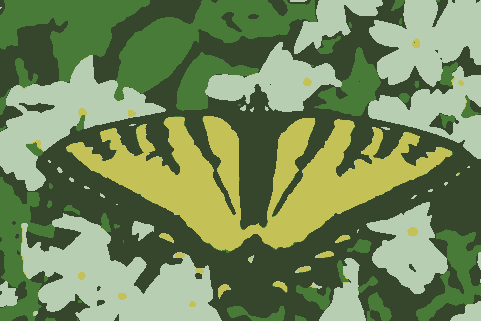}}\\[0.15cm]
\subfloat[Phase 1, LCV]
 {\includegraphics[width=5.00cm]{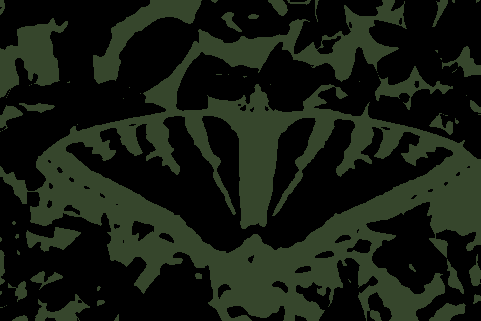}}\\
\subfloat[Phase 2, LCV\label{fig:lcv_phase2}]
 {\includegraphics[width=5.00cm]{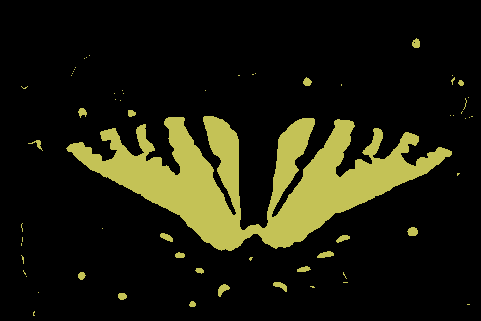}}\\
\subfloat[Phase 3, LCV\label{fig:lcv_phase3}]
 {\includegraphics[width=5.00cm]{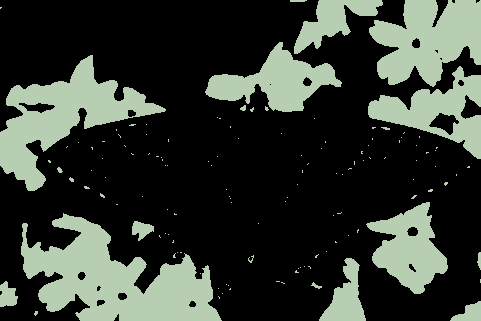}}\\
\subfloat[Phase 4, LCV\label{fig:lcv_phase4}]
 {\includegraphics[width=5.00cm]{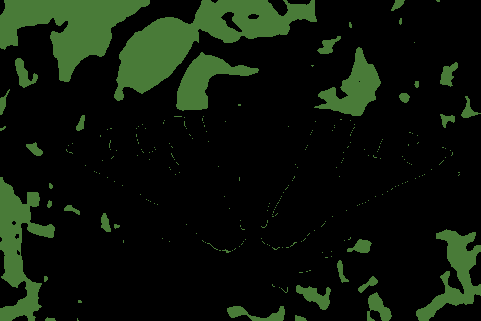}}
\end{tabular}
\\
\end{tabular}

\caption{Four-phase segmentation results for a natural color image exhibiting intensity inhomogeneity along with the individual phases. 
Parameters are $\lambda = 40$ and $dt = 60$ for both models and $\beta = 60$ for LCV.   
The LCV model successfully separates the yellow region of the wings from the white region of the flowers as two phases, whereas CV merges these regions in one phase because of global intensity averaging.}
\label{fig:4phase-butterfly}
\end{figure}

Finally, we consider four-phase color segmentation on more complex natural scenes, focusing on two representative images. Figures~\ref{fig:4phase-church} and~\ref{fig:4phase-butterfly} compare the results obtained with the CV and LCV models. While both approaches produce similar coarse segmentations, the LCV model consistently achieves a more faithful separation of neighboring regions that exhibit subtle color differences.

We first apply a color version of Algorithm~\ref{alg:fourphase} to a natural image of a church, shown in Figure~\ref{fig:church_original}. The corresponding four-phase segmentation results are presented in Figure~\ref{fig:4phase-church}, using $\lambda=40$ and $dt=50$ for both models and $\beta=200$ only for LCV. Although CV and LCV yield comparable coarse segmentations, LCV provides a noticeably finer discrimination of the ladder and railing structures near the entrance, preventing their artificial merging caused by local illumination variations. 

For the butterfly image in Figure~\ref{fig:orig_butterfly}, we use the same algorithm with $\lambda=40$ and $dt=60$ for both models and $\beta=60$ only for LCV. The corresponding results are shown in Figure~\ref{fig:4phase-butterfly}. Compared with CV, the LCV model produces a segmentation that is more faithful to the original image, specifically by separating the yellow region of the wings from the white region of the flowers. This distinction is clearly visible in the LCV phases shown in Figures~\ref{fig:lcv_phase2} and~\ref{fig:lcv_phase3}. By contrast, likely because of intensity inhomogeneity, the CV model merges these regions into a single phase, as seen in Figure~\ref{fig:cv_phase2}. 
Overall, these experiments demonstrate that the proposed MBO-based LCV framework extends naturally to multiphase color segmentation and remains robust in the presence of local intensity inhomogeneities and nonuniform lighting conditions.

\section{Conclusion}\label{sec:conclusion}

In this work, we presented an efficient implementation of the LCV model based on the MBO scheme. Starting from the continuous variational formulation, the algorithm consists of explicitly solving the heat equations and ODEs followed by thresholding. This structure leads to a numerically stable method with low computational complexity and straightforward implementation.\smallskip

A broad range of numerical experiments demonstrates the versatility of the proposed approach. The method was successfully applied to two-phase and multiphase segmentation problems on grayscale and color images, including medical imaging data, microscopy images, and natural scenes. In all tested cases, the algorithm produces sharp and stable interfaces and exhibits robust behavior in the presence of intensity inhomogeneities. Comparative results with the CV model show that incorporating local fidelity terms significantly improves segmentation accuracy when global piecewise-constant assumptions are violated. The MBO-based formulation preserves the advantages of the LCV model while providing a fast solver that scales efficiently to large images and multiple phases.\smallskip

Owing to its simplicity, efficiency, and robustness, the proposed method constitutes a practical and reproducible tool for variational image segmentation. The implementation is particularly well suited for experimental analysis and serves as a flexible framework for extending local variational models to more complex segmentation tasks.

\section*{Acknowledgments}
The authors would like to thank the reviewers for their helpful feedback and comments to help improve the presentation of the manuscript. In addition, we thank Dr. Marina Gardella for implementing the online demo for our paper. 

\section*{Image Credits}\small

\noindent
\includegraphics[height=2em]{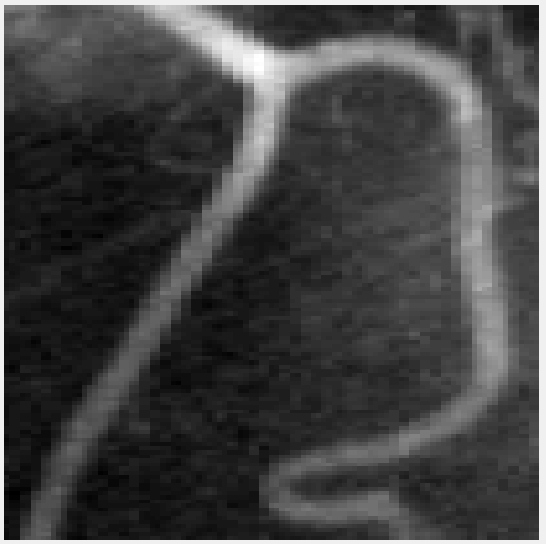}\,
\includegraphics[height=2em]{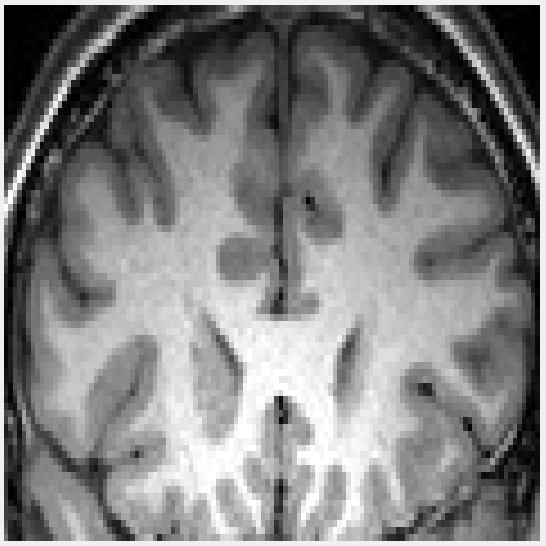}\,
Images adapted from Li et al.~\cite{li2008minimization}, available at \\ 
\url{http://www.imagecomputing.org/~cmli/code/}.\\[0.4em]

\noindent
\includegraphics[height=2em]{images/turtle/turtle_rgb.pdf}\,
\includegraphics[height=2em]{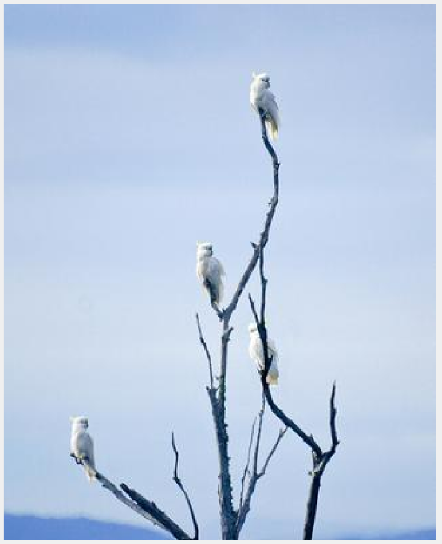}\,
Images from the DUTS Salient Object Detection Dataset~\cite{wang2017learning}, \\ 
\url{http://saliencydetection.net/duts/}.\\[0.4em]

\noindent
\includegraphics[height=2em]{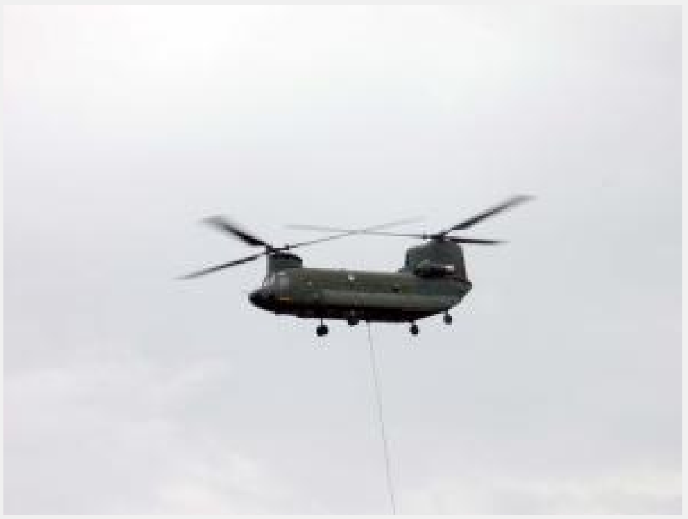}\,
Image from the Weizmann Segmentation Evaluation Database~\cite{AlpertGBB07},\\  
\url{http://www.wisdom.weizmann.ac.il/~vision/Seg_Evaluation_DB/}.\\[0.4em]

\noindent
\includegraphics[height=2em]{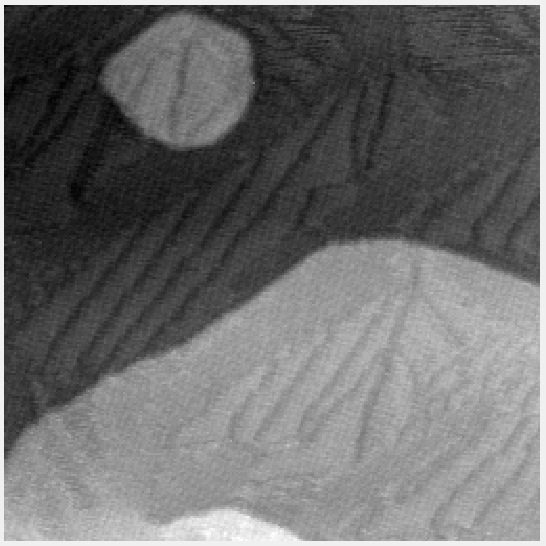}\,
\includegraphics[height=2em]{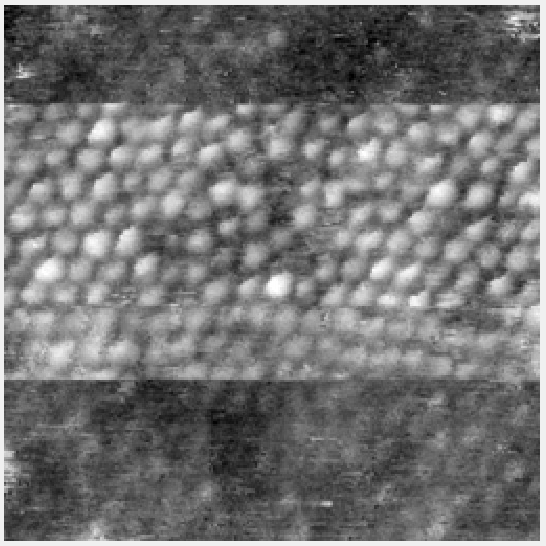}\,
Microscopy images courtesy of D.~Goronzy, A.~Guttentag, and P.~Weiss\\  
(California NanoSystems Institute, UCLA).\\[0.4em]

\noindent
\includegraphics[height=2em]{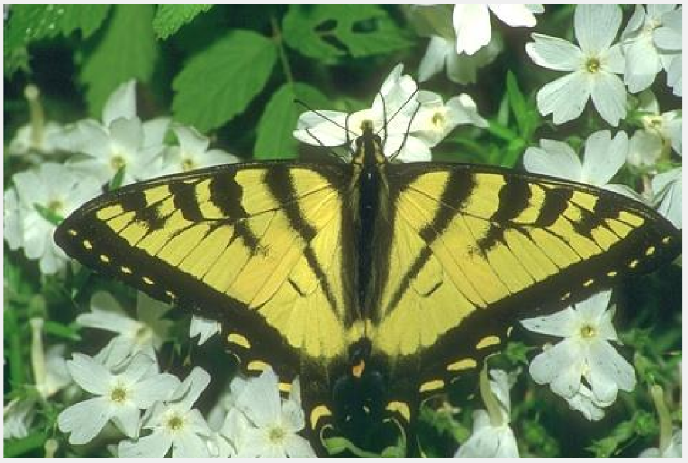}\,
\includegraphics[height=2em]{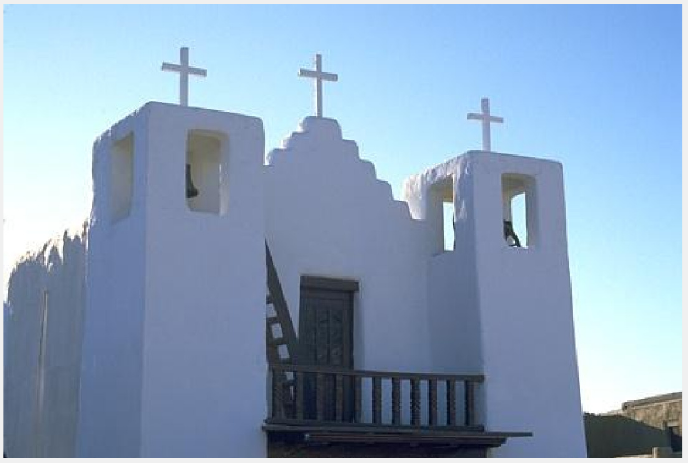}\,
Images from the Berkeley Segmentation Dataset~\cite{martin2001database},\\ 
\url{https://www2.eecs.berkeley.edu/Research/Projects/CS/vision/bsds/}.\\[0.4em]

\noindent
\includegraphics[height=2em]{images_python/map_image.png}\,
Image provided by Blaise Mibeck.\\[0.4em]

\noindent
\includegraphics[height=2em]{images_python/manuscript_image.png}\,
Image provided by Biblioteca Nacional de España.

\bibliographystyle{siam}      
\small
\bibliography{article} 

@article{ali2016variational,
  title={A variational model with hybrid images data fitting energies for segmentation of images with intensity inhomogeneity},
  author={Ali, Haider and Badshah, Noor and Chen, Ke and Khan, Gulzar Ali},
  journal={Pattern Recognition},
  volume={51},
  pages={27--42},
  year={2016},
  publisher={Elsevier}
}

@inproceedings{AlpertGBB07,
   author = {Sharon Alpert and Meirav Galun and Ronen Basri and Achi Brandt},
   title = {Image Segmentation by Probabilistic Bottom-Up Aggregation and Cue                             Integration.},
   booktitle = {Proceedings of the IEEE Conference on Computer Vision and Pattern                                 Recognition},
   month = {June},
   year = {2007}
}

@article{bezdek1993review,
  title={Review of {MR} image segmentation techniques using pattern recognition},
  author={Bezdek, James C and Hall, LO and Clarke, LP},
  journal={Medical Physics},
  volume={20},
  number={4},
  pages={1033--1048},
  year={1993},
  publisher={Wiley Online Library}
}

@article{boyd2011distributed,
  title={Distributed optimization and statistical learning via the alternating direction method of multipliers},
  author={Boyd, Stephen and Parikh, Neal and Chu, Eric and Peleato, Borja and Eckstein, Jonathan and others},
  journal={Foundations and Trends{\textregistered} in Machine learning},
  volume={3},
  number={1},
  pages={1--122},
  year={2011},
  publisher={Now Publishers, Inc.}
}

@article{brown2012completely,
  title={Completely convex formulation of the {C}han-{V}ese image segmentation model},
  author={Brown, Ethan S and Chan, Tony F and Bresson, Xavier},
  journal={International Journal of Computer Vision},
  volume={98},
  number={1},
  pages={103--121},
  year={2012},
  publisher={Springer}
}

@article{bui2024efficient,
  title={An efficient smoothing and thresholding image segmentation framework with weighted anisotropic-isotropic total variation},
  author={Bui, Kevin and Lou, Yifei and Park, Fredrick and Xin, Jack},
  journal={Communications on Applied Mathematics and Computation},
  volume={6},
  number={2},
  pages={1369--1405},
  year={2024},
  publisher={Springer}
}

@article{bui2019segmentation,
  title={Segmentation of scanning tunneling microscopy images using variational methods and empirical wavelets},
  author={Bui, Kevin and Fauman, Jacob and Kes, David and Mandiola, Leticia Torres and Ciomaga, Adina and Salazar, Ricardo and Bertozzi, Andrea L and Gilles, J{\'e}r{\^o}me and Goronzy, Dominic P and Guttentag, Andrew I and others},
  journal={Pattern Analysis and Applications},
  pages={1--27},
  year={2019},
  publisher={Springer}
}

@article{cai2017three,
  title={A three-stage approach for segmenting degraded color images: Smoothing, lifting and thresholding ({SLaT})},
  author={Cai, Xiaohao and Chan, Raymond and Nikolova, Mila and Zeng, Tieyong},
  journal={Journal of Scientific Computing},
  volume={72},
  pages={1313--1332},
  year={2017},
  publisher={Springer}
}

@article{caselles1997geodesic,
  title={Geodesic active contours},
  author={Caselles, Vicent and Kimmel, Ron and Sapiro, Guillermo},
  journal={International Journal of Computer Vision},
  volume={22},
  number={1},
  pages={61--79},
  year={1997},
  publisher={Springer}
}

@article{chan2000active,
  title={Active contours without edges for vector-valued images},
  author={Chan, Tony F and Sandberg, B Yezrielev and Vese, Luminita A},
  journal={Journal of Visual Communication and Image Representation},
  volume={11},
  number={2},
  pages={130--141},
  year={2000},
  publisher={Citeseer}
}

@article{chan-vese-2001,
  title={Active contours without edges},
  author={Chan, Tony F and Vese, Luminita A},
  journal={IEEE Transactions on Image Processing},
  volume={10},
  number={2},
  pages={266--277},
  year={2001},
  publisher={IEEE}
}

@article{chambolle2011first,
  title={A first-order primal-dual algorithm for convex problems with applications to imaging},
  author={Chambolle, Antonin and Pock, Thomas},
  journal={Journal of Mathematical Imaging and Vision},
  volume={40},
  number={1},
  pages={120--145},
  year={2011},
  publisher={Springer}
}

@article{chan2006algorithms,
  title={Algorithms for finding global minimizers of image segmentation and denoising models},
  author={Chan, Tony F and Esedoglu, Selim and Nikolova, Mila},
  journal={SIAM Journal on Applied Mathematics},
  volume={66},
  number={5},
  pages={1632--1648},
  year={2006},
  publisher={SIAM}
}

@article{cohen1991active,
  title={On active contour models and balloons},
  author={Cohen, Laurent D},
  journal={CVGIP: Image Understanding},
  volume={53},
  number={2},
  pages={211--218},
  year={1991},
  publisher={Elsevier}
}

@article{dong2013new,
  title={A new level set method for inhomogeneous image segmentation},
  author={Dong, Fangfang and Chen, Zengsi and Wang, Jinwei},
  journal={Image and Vision Computing},
  volume={31},
  number={10},
  pages={809--822},
  year={2013},
  publisher={Elsevier}
}

@article{esedog2006threshold,
  title={Threshold dynamics for the piecewise constant {M}umford--{S}hah functional},
  author={Esedog, Selim and Tsai, Yen-Hsi Richard and others},
  journal={Journal of Computational Physics},
  volume={211},
  number={1},
  pages={367--384},
  year={2006},
  publisher={Elsevier}
}

@article{getreuer2012chan,
  title={Chan-{V}ese segmentation},
  author={Getreuer, Pascal},
  journal={Image Processing On Line},
  volume={2},
  pages={214--224},
  year={2012}
}

@article{guttentag2016hexagons,
  title={Hexagons to ribbons: Flipping cyanide on {Au} $\{$111$\}$},
  author={Guttentag, Andrew I and Barr, Kristopher K and Song, Tze-Bin and Bui, Kevin V and Fauman, Jacob N and Torres, Leticia F and Kes, David D and Ciomaga, Adina and Gilles, J{\'e}r{\^o}me and Sullivan, Nichole F and others},
  journal={Journal of the American Chemical Society},
  volume={138},
  number={48},
  pages={15580--15586},
  year={2016},
  publisher={ACS Publications}
}

@article{kass1988snakes,
  title={Snakes: Active contour models},
  author={Kass, Michael and Witkin, Andrew and Terzopoulos, Demetri},
  journal={International Journal of Computer Vision},
  volume={1},
  number={4},
  pages={321--331},
  year={1988},
  publisher={Springer}
}

@article{li2008minimization,
  title={Minimization of region-scalable fitting energy for image segmentation},
  author={Li, Chunming and Kao, Chiu-Yen and Gore, John C and Ding, Zhaohua},
  journal={IEEE Transactions on Image Processing},
  volume={17},
  number={10},
  pages={1940--1949},
  year={2008},
  publisher={IEEE}
}

@article{li2010multiphase,
  title={A multiphase image segmentation method based on fuzzy region competition},
  author={Li, Fang and Ng, Michael K and Zeng, Tie Yong and Shen, Chunli},
  journal={SIAM Journal on Imaging Sciences},
  volume={3},
  number={3},
  pages={277--299},
  year={2010},
  publisher={SIAM}
}

@inproceedings{martin2001database,
  title={A database of human segmented natural images and its application to evaluating segmentation algorithms and measuring ecological statistics},
  author={Martin, David and Fowlkes, Charless and Tal, Doron and Malik, Jitendra},
  booktitle={Proceedings Eighth IEEE International Conference on Computer Vision. ICCV 2001},
  volume={2},
  pages={416--423},
  year={2001},
  organization={IEEE}
}

@article{merriman1994motion,
  title={Motion of multiple junctions: A level set approach},
  author={Merriman, Barry and Bence, James K and Osher, Stanley J},
  journal={Journal of Computational Physics},
  volume={112},
  number={2},
  pages={334--363},
  year={1994},
  publisher={Elsevier}
}

@article{modica1987gradient,
  title={The gradient theory of phase transitions and the minimal interface criterion},
  author={Modica, Luciano},
  journal={Archive for Rational Mechanics and Analysis},
  volume={98},
  number={2},
  pages={123--142},
  year={1987},
  publisher={Springer-Verlag}
}

@article{mumford1989optimal,
  title={Optimal approximations by piecewise smooth functions and associated variational problems},
  author={Mumford, David and Shah, Jayant},
  journal={Communications on pure and applied mathematics},
  volume={42},
  number={5},
  pages={577--685},
  year={1989},
  publisher={Wiley Online Library}
}

@article{osher1988fronts,
  title={Fronts propagating with curvature-dependent speed: algorithms based on {H}amilton-{J}acobi formulations},
  author={Osher, Stanley and Sethian, James A},
  journal={Journal of 
  Computational Physics},
  volume={79},
  number={1},
  pages={12--49},
  year={1988},
  publisher={Elsevier}
}

@article{paschos2001perceptually,
  title={Perceptually uniform color spaces for color texture analysis: an empirical evaluation},
  author={Paschos, George},
  journal={IEEE transactions on Image Processing},
  volume={10},
  number={6},
  pages={932--937},
  year={2001},
  publisher={IEEE}
}

@article{pierre2014segmentation,
  title={Segmentation with Active Contours},
  author={Pierre, Fabien and Amendola, Mathieu and Bigeard, Cl{\'e}mence and Ruel, Timoth{\'e} and Villard, Pierre-Fr{\'e}d{\'e}ric},
  year={2014},
  journal={Image Processing On Line}
}

@article{thomas2018acid,
  title={Acid--Base Control of Valency within Carboranedithiol Self-Assembled Monolayers: Molecules Do the Can-Can},
  author={Thomas, John C and Goronzy, Dominic P and Serino, Andrew C and Auluck, Harsharn S and Irving, Olivia R and Jimenez-Izal, Elisa and Deirmenjian, Jacqueline M and Machacek, Jan and Sautet, Philippe and Alexandrova, Anastassia N and others},
  journal={ACS Nano},
  volume={12},
  number={3},
  pages={2211--2221},
  year={2018},
  publisher={ACS Publications}
}

@article{vese2002multiphase,
  title={A multiphase level set framework for image segmentation using the {M}umford and {S}hah model},
  author={Vese, Luminita A and Chan, Tony F},
  journal={International Journal of Computer Vision},
  volume={50},
  number={3},
  pages={271--293},
  year={2002},
  publisher={Springer}
}

@article{wang2010efficient,
  title={An efficient local {C}han--{V}ese model for image segmentation},
  author={Wang, Xiao-Feng and Huang, De-Shuang and Xu, Huan},
  journal={Pattern Recognition},
  volume={43},
  number={3},
  pages={603--618},
  year={2010},
  publisher={Elsevier}
}

@article{wang2013efficient,
  title={An efficient operator splitting method for local region {C}han-{V}ese model},
  author={Wang, Hui and Huang, Ting-Zhu and Liu, Jun},
  journal={EURASIP Journal on Advances in Signal Processing},
  volume={2013},
  number={1},
  pages={97},
  year={2013},
  publisher={Springer}
}

@inproceedings{wang2017learning,
  title={Learning to detect salient objects with image-level supervision},
  author={Wang, Lijun and Lu, Huchuan and Wang, Yifan and Feng, Mengyang and Wang, Dong and Yin, Baocai and Ruan, Xiang},
  booktitle={Proceedings of the IEEE Conference on Computer Vision and Pattern Recognition},
  pages={136--145},
  year={2017}
}

@article{zosso2017image,
  title={Image segmentation with dynamic artifacts detection and bias correction},
  author={Zosso, Dominique and An, Jing and Stevick, James and Takaki, Nicholas and Weiss, Morgan and Slaughter, Liane S and Cao, Huan H and Weiss, Paul S and Bertozzi, Andrea L},
  journal={Inverse Problems and Imaging},
  volume={11},
  number={3},
  pages={577--600},
  year={2017}
}

\end{document}